\documentclass[onefignum,onetabnum]{siamonline171218}

\usepackage{latexsym,amssymb,amsmath,amsbsy,amsopn,amstext,amsxtra,color,bm,calc,ifpdf}
\usepackage{float}
\usepackage{enumerate}
\usepackage{fancyhdr}
\usepackage{listings}
\usepackage{multicol}
\usepackage{multirow}
\usepackage{makeidx}
\usepackage{makecell}
\usepackage{xcolor} 
\usepackage{graphicx}
\usepackage{subfig}
\usepackage{appendix}
\usepackage{booktabs}

\newcommand {\Figref}[1]{Figure \ref{#1}}

\newcommand {\Romnum}[1]{\uppercase\expandafter{\romannumeral #1}}

\usepackage{tikz}
\newcommand*\circled[1]{\tikz[baseline=(char.base)]{
		\node[shape=circle,draw,inner sep=0.6pt] (char) {#1};}}
\usepackage{ulem}

\usepackage{lipsum}
\usepackage{amsfonts}
\usepackage{graphicx}
\usepackage{epstopdf}
\usepackage{algorithmic}
\ifpdf
  \DeclareGraphicsExtensions{.eps,.pdf,.png,.jpg}
\else
  \DeclareGraphicsExtensions{.eps}
\fi

\usepackage{enumitem}
\setlist[enumerate]{leftmargin=.5in}
\setlist[itemize]{leftmargin=.5in}

\newsiamremark{remark}{Remark}
\newsiamremark{hypothesis}{Hypothesis}
\crefname{hypothesis}{Hypothesis}{Hypotheses}
\newsiamthm{claim}{Claim}

\headers{Local transfer learning GP}{Wang et al.}

\title{Local transfer learning Gaussian process modeling, with applications to surrogate modeling of expensive computer simulators
}

\author{Xinming Wang\thanks{Department of Industrial Engineering and Management, Peking University}
\and Simon Mak\thanks{Department of Statistical Science, Duke University (\email{sm769@duke.edu})}
\and John Miller\thanks{Department of Statistical Science, Duke University}
\and Jianguo Wu\thanks{Department of Industrial Engineering and Management, Peking University (\email{j.wu@pku.edu.cn})}
\funding{The authors gratefully acknowledge support from NSF CSSI 2004571, NSF DMS 2210729, NSF DMS 2316012 and DE-SC0024477.}
}

\usepackage{amsopn}

\makeatletter
\newcommand*{\addFileDependency}[1]{
  \typeout{(#1)}
  \@addtofilelist{#1}
  \IfFileExists{#1}{}{\typeout{No file #1.}}
}
\makeatother

\ifpdf
\hypersetup{
	pdftitle={An Example Article},
	pdfauthor={D. Doe, P. T. Frank, and J. E. Smith}
}
\fi

\usepackage{hyperref}  

\usepackage{pdfpages}

\begin{document}
	
	\maketitle
	
	\begin{abstract}
		A critical bottleneck for scientific progress is the costly nature of computer simulations for complex systems. Surrogate models provide an appealing solution: such models are trained on simulator evaluations, then used to emulate and quantify uncertainty on the expensive simulator at unexplored inputs. In many applications, one often has available data on related systems. For example, in designing a new jet turbine, there may be existing studies on turbines with similar configurations. A key question is how information from such ``source'' systems can be transferred for effective surrogate training on the ``target'' system of interest. We thus propose a new LOcal transfer Learning Gaussian Process (LOL-GP) model, which leverages a carefully-designed Gaussian process to transfer such information for surrogate modeling. The key novelty of the LOL-GP is a latent regularization model, which identifies regions where transfer should be performed and regions where it should be avoided. Such a ``local transfer'' property is present in many scientific systems: at certain parameters, systems may behave similarly and thus transfer is beneficial; at other parameters, they may behave differently and thus transfer is detrimental. By accounting for local transfer, the LOL-GP can temper the risk of ``negative transfer'', i.e., the risk of worsening predictive performance from information transfer. We derive a Gibbs sampling algorithm for efficient posterior predictive sampling on the LOL-GP, for both the multi-source and multi-fidelity transfer settings. We then show, via a suite of numerical experiments and an application for jet turbine design, the improved surrogate performance of the LOL-GP over existing methods.
		
	\end{abstract}
	
	\begin{keywords}
		Computer experiments, Gaussian processes, regularization, surrogate modeling, transfer learning.
	\end{keywords}
	
	\begin{AMS}
		68Q25, 68R10, 68U05
	\end{AMS}
	
	\section{Introduction} 
	\label{sec: intro}
	With fundamental advances in scientific modeling and computing, computer simulations are becoming an increasingly valuable tool for investigating complex physical phenomena. Such virtual simulations have been successfully employed in important scientific and engineering applications, including rocket design \cite{Rocket2018}, climate science \cite{Climate2014} and high-energy physics \cite{li2023additive}. One critical bottleneck, however, is that these simulations are typically computationally costly. For complex phenomena such as rocket propulsion, the evaluation of the simulator $f(\bm{x})$ at an input $\bm{x}$ can easily require millions of CPU hours \cite{yeh2018common}. An effective solution is \textit{surrogate modeling} \cite{gramacy2020surrogates}: one first performs simulations on a carefully selected set of design points, then uses the simulated data to train a predictive model that efficiently ``emulates'' and quantifies uncertainty on the expensive simulator. In what follows, we focus on \textit{probabilistic} surrogates that provide uncertainty quantification (UQ) on its predictions, as such probabilistic UQ is essential for many facets of scientific decision-making \cite{gramacy2020surrogates}.
	
	A popular probabilistic surrogate model is the Gaussian process (GP; \cite{GP2006}), a flexible Bayesian nonparametric model widely used in statistics and machine learning. The primary appeal of GPs as surrogates is that, conditioned on simulated training data, its posterior predictive distribution can be evaluated as a closed-form expression. Such an expression not only permits efficient prediction and UQ on the expensive simulator $f(\cdot)$, but also facilitates effective use of the trained surrogate for downstream objectives, including calibration \cite{cao2021determining,everett2021multisystem, plumlee2017bayesian}, optimization \cite{miller2024targeted, chen2023hierarchical}, sequential design \cite{sauer2023active,chen2022adaptive} and control \cite{narayanan2024misfire}. In recent years, there has been much work on improving the expressiveness of GPs for broad applications, including deep GPs \cite{sauer2023active}, neural network GPs \cite{lee2017deep} and shrinkage GPs \cite{tang2023hierarchical}.
	
	Despite this work, an inherent bottleneck for surrogate modeling is the expensive (thus limited) nature of simulation training data. This is exacerbated by the reality that, in scientific applications, systems may have high-dimensional input spaces that require large training sample sizes for accurate surrogates. A recent development in machine learning, called \textit{transfer learning} \cite{TransferSurvey2020}, offers a promising solution. The key idea is to ``transfer'' knowledge from \textit{source} systems with ample amounts of data, to \textit{target} systems with limited data. For surrogates, the target system is the simulator of interest, and the source systems may be simulators from related prior studies. Despite successful applications of transfer learning in machine learning (e.g., text classification \cite{do2005transfer} and spam filtering \cite{bickel2006ecml}), there is scant literature on its broad use for probabilistic surrogate modeling. \cite{TransferSurrogate-HeavyIon2022} proposed an extension of the Kennedy-O'Hagan (KO) model \cite{KO2000} for transfer learning of GP surrogates, which we investigate later.
	While the general use of transfer learning for surrogates may be limited, a specific instance has been explored under the topic of multi-fidelity learning. Using transfer learning terminology, multi-fidelity learning aims to transfer knowledge from cheaper low-fidelity simulators (i.e., source systems) to emulate a costly high-fidelity simulator (i.e., target system). Beginning with the seminal KO model \cite{KO2000}, there has been considerable work on multi-fidelity surrogates for scientific applications, including \cite{BayesianKO2008,Gratiet2013,tuo2014surrogate,NonlinearGPMF2017,ji2022conglomerate,ji2023graphical}.
	
	A critical challenge for transfer learning is the potential of \textit{negative transfer} \cite{TransferSurvey2020}, where the transfer of information can considerably \textit{worsen} performance on the target system. Such negative transfer can be disastrous for surrogate training, resulting in highly inaccurate surrogates with unreliable UQ. As we show later, similar issues can arise for existing multi-fidelity surrogates, where the use of low-fidelity data may worsen predictions on the high-fidelity simulator. To mitigate this, recent machine learning works have explored selection criteria for choosing which sources to include for transfer learning \cite{JSTransfer2021,ActiveForget2021, RegularizedMGP2022, OnNegativeTransfer2022}. For physical science applications, however, the key consideration is typically not \textit{which} sources to select, but rather \textit{how} transfer is modeled. Consider, for example, our motivating application (see Section \ref{sec: stress emulation}) on the stress analysis of a jet engine turbine blade. It is known \cite{salwan2021comparison} that different blade materials induce similar stress profiles at certain operating conditions, but considerably different profiles at other conditions. Transfer learning can thus be beneficial in the first case, but detrimental in the second. This highlights a so-called ``local transfer'' property, where systems may be related at certain parametrizations (e.g., due to similar dominant physics), but considerably different at other parametrizations (e.g., when such physics are less pronounced); \cite{TransferSurrogate-HeavyIon2022} provides an example of this in high-energy physics. Local transfer is also expected in many multi-fidelity systems. For example, in our jet turbine application, the simulated stress profiles at low and high fidelity may be similar at certain operating conditions, but considerably different at others. Incorporating such local transfer can help temper the risk of negative transfer by learning appropriate regions for transfer; existing methods, however, do not account for this. 
	
	We thus propose a new LOcal transfer Learning GP (LOL-GP), which models for local transfer via a flexible Bayesian nonparametric modeling framework. The key novelty of the LOL-GP is a latent regularization model, which identifies from data regions where transfer should be performed and regions where it should be avoided. By modeling such a local transfer property, the LOL-GP can trade off between simple and complex transfer models, enabling robust and reliable transfer learning for surrogate modeling. We derive an efficient Gibbs sampler for posterior predictive sampling from the LOL-GP, for both the multi-source and multi-fidelity transfer settings. We then show, in a suite of numerical experiments and an application to jet turbine design, the improved surrogate performance of the LOL-GP over existing methods, particularly in tempering the risks of negative transfer.
	
	The paper is organized as follows. Section \ref{sec: background} provides background on GP modeling and investigates limitations of existing transfer models. Section \ref{sec: model} presents the proposed LOL-GP and its adaptation for multi-fidelity learning. Section \ref{sec: fitting and prediction} details the workflow for model training and prediction. Sections \ref{sec: numerical experiments} and \ref{sec: stress emulation} compare the effectiveness of the proposed model with the state-of-the-art in a suite of numerical experiments and for our jet turbine application. Section \ref{sec:con} concludes the paper.

	\section{Background and Motivation}
	\label{sec: background}
	We first provide a brief overview of GPs and related work on transfer learning surrogates, then present a motivating example outlining the risk of negative transfer for such models.
	
	\subsection{Gaussian process modeling}
	\label{subsec: gp model}
	Let $\bm{x}$ be the simulation inputs on the input space $\mathbb{X} = [0,1]^d$ (i.e., all inputs are normalized to be on a unit hypercube; see \cite{gramacy2020surrogates}), and let $f(\bm{x}) \in \mathbb{R}$ be the simulated output at inputs $\bm{x}$. A (zero-mean) GP places the following prior stochastic process on the black-box function $f(\cdot)$:
	\begin{align}
		f(\cdot) \sim \mathcal{GP}\{0, k(\cdot,\cdot)\}.
	\end{align}
	Here, $k(\cdot,\cdot): \mathbb{X} \times \mathbb{X} \rightarrow \mathbb{R}$ is its covariance function, which dictates the smoothness of sample paths from the GP. Popular choices for $k$ include the squared-exponential and Mat\'ern-$\nu$ kernels, which induce infinitely-differentiable and $(\lfloor \nu \rfloor-1)$-differentiable sample paths from the GP, respectively (see \cite{santner2003design}). 
	
	Suppose we then run the expensive simulator at design points $\mathcal{X} = \{\bm{x}_1, \cdots, \bm{x}_n\}$, yielding outputs $\bm{f} = [{f}(\bm{x}_1), \cdots, {f}(\bm{x}_n)]$. In what follows, we presume the simulator is deterministic, in that the same output $f(\bm{x})$ is observed every time the simulator is run at inputs $\bm{x}$; this can be extended in a straight-forward manner for simulators with Gaussian noise (see \cite{ankenman2008stochastic}). Suppose we wish to predict $f$ at a new input $\bm{x}_{\rm new}$. Conditioning on training data $\bm{f}$, the posterior predictive distribution of $f(\bm{x}_{\rm new})$ can be shown to be Gaussian, i.e.:
	\begin{align}
		\begin{split}
			f(\bm{x}_{\rm new}) | \bm{f} \sim \mathcal{N}\{\mu(\bm{x}_{\rm new}), \sigma^2(\bm{x}_{\rm new})\}
			\label{eq: gp prediction}
		\end{split}
	\end{align}
	where its mean and variance are given by the closed-form expressions:
	\begin{align}
		\begin{split}
			\mu(\bm{x}_{\rm new}) &= \bm{k}^T(\mathcal{X}, \bm{x}_{\rm new}) [\bm{K}(\mathcal{X}, \mathcal{X})]^{-1} \bm{f},\\
			\sigma^2(\bm{x}_{\rm new}) &= {k}(\bm{x}_{\rm new},\bm{x}_{\rm new}) - \bm{k}^T(\mathcal{X}, \bm{x}_{\rm new}) [\bm{K}(\mathcal{X}, \mathcal{X})]^{-1}\bm{k}(\mathcal{X}, \bm{x}_{\rm new}).
		\end{split}
		\label{eq: gp prediction2}
	\end{align}
	Here, $\bm{K}(\mathcal{X}, \mathcal{X}) = [k(\bm{x},\bm{x}')]_{\bm{x},\bm{x}' \in \mathcal{X}}$ is the covariance matrix at the simulated inputs in $\mathcal{X}$, and $\bm{k}(\mathcal{X}, \bm{x}_{\rm new}) = [k(\bm{x}, \bm{x}_{\rm new})]_{\bm{x} \in \mathcal{X}}$ is the covariance vector between the simulated inputs $\mathcal{X}$ and the new point $\bm{x}_{\rm new}$. 
	
	\subsection{Existing transfer learning surrogates}
	\label{subsec: surrogates}
	Next, we outline existing work on probabilistic surrogates that leverage transfer learning. As mentioned earlier, much of this work focuses specifically on multi-fidelity surrogates. An early seminal work on multi-fidelity surrogates is the Kennedy-O'Hagan model \cite{KO2000}, which aims to predict a high-fidelity simulator $f_T(\cdot)$ by borrowing information from a sequence of $L \geq 1$ lower fidelity simulators $f_{L}(\cdot), \cdots, f_{1}(\cdot)$, ranked by decreasing fidelities. To do this, the KO model presumes:
	\begin{align}
		\begin{split}
			f_{l+1}(\bm{x}) &= \rho f_l(\bm{x}) + \delta_l(\bm{x}), \quad l = 1, \cdots, L,
			\label{eq: KO}
		\end{split}
	\end{align}
	where $f_{L+1}(\cdot) := f_T(\cdot)$, and $f_1(\cdot)$ and $\{\delta_l(\cdot)\}_{l=1}^L$ follow independent GP priors.
	Here, $\rho$ can be viewed as a ``transfer'' parameter that facilitates information transfer over fidelity levels, and $\delta_l(\bm{x})$ models the ``discrepancy'' between fidelity levels $l$ and $l+1$. There has been considerable extensions of the KO model, including with mesh densities \cite{tuo2014surrogate, ji2022conglomerate}, graphical models \cite{ji2023graphical} and for Bayesian optimization \cite{konomi2021bayesian}. A recent work \cite{TransferSurrogate-HeavyIon2022} adopted the KO model for transfer learning of surrogates in high-energy physics; here, transfer is from a related source system to the target system, where both systems are not related via fidelity. Letting $f_S(\cdot)$ and $f_T(\cdot)$ be the source and target simulators, respectively, the model in \cite{TransferSurrogate-HeavyIon2022} takes the form:
	\begin{equation}
		f_T(\bm{x}) = \rho f_S(\bm{x}) + \delta(\bm{x}),
		\label{eq: KOtr}
	\end{equation}
	where $f_S(\cdot)$ and $\delta(\cdot)$ follow independent GP priors. Such a model was shown to be effective for transfer learning between different particle collision systems for surrogate modeling \cite{TransferSurrogate-HeavyIon2022}.
	
	A potential limitation of the KO model \eqref{eq: KO} (and similarly of model \eqref{eq: KOtr}) is that the modeled transfer is too simplistic, which may inhibit transfer learning and even induce \textit{negative} transfer. For the multi-fidelity setting, \cite{BayesianKO2008} proposed the following model extending \eqref{eq: KO}:
	\begin{align}
		\begin{split}
			f_{l+1}(\bm{x}) &= \rho_l(\bm{x}) f_{l}(\bm{x}) + \delta_{l}(\bm{x}), \quad l = 1, \cdots, L.
			\label{eq: BKO}
		\end{split}
	\end{align}
	Here, $\{\rho_l(\bm{x})\}_l$ follow independent GPs, which facilitates a more flexible \textit{input-dependent} transfer of information from low to high fidelity. Under this so-called Bayesian KO (BKO) model, \cite{BayesianKO2008} introduced an efficient Gibbs sampler for sampling the posterior predictive distribution of the high-fidelity code $f_T(\cdot) = f_{L+1}(\cdot)$. A similar extension can be used on \eqref{eq: KOtr} for transfer learning; we will investigate this later. While such an extension indeed offers greater flexibility for the transfer model, this may be a ``double-edged sword'': the transfer model may be prone to overfitting with limited data, which in turn may induce negative transfer. We shall see the potential risk at each end of this trade-off in the motivating example later.
	
	Finally, beyond GPs, there is recent literature on transfer learning using neural network surrogates. This includes the works of \cite{TransferSurrogate-MetaLearning2022, song2022transfer,TransferSurrogate-TemperatureField2021}, which employ a pre-training of the network using source data, then a fine-tuning of such a model using target data. Recent work has explored transfer learning within neural network surrogates for thermodynamics \cite{TransferSurrogate-TemperatureField2021}, reliability \cite{TransferSurrogate-MetaLearning2022}, and inertial confinement fusion applications \cite{Humbird2020, Humbird2022, VanderWal2023}. However, as mentioned in the Introduction, existing surrogates of this flavor are largely deterministic in nature, and do not provide the probabilistic UQ required in our applications. Furthermore, such transfer models may similarly be overly flexible and may experience negative transfer with limited target data. We will show this later in numerical experiments.
	
	\subsection{A motivating example}
	\label{subsec: motivation}
\begin{figure}[!t]
		\centering
		\includegraphics[width=0.40\linewidth]{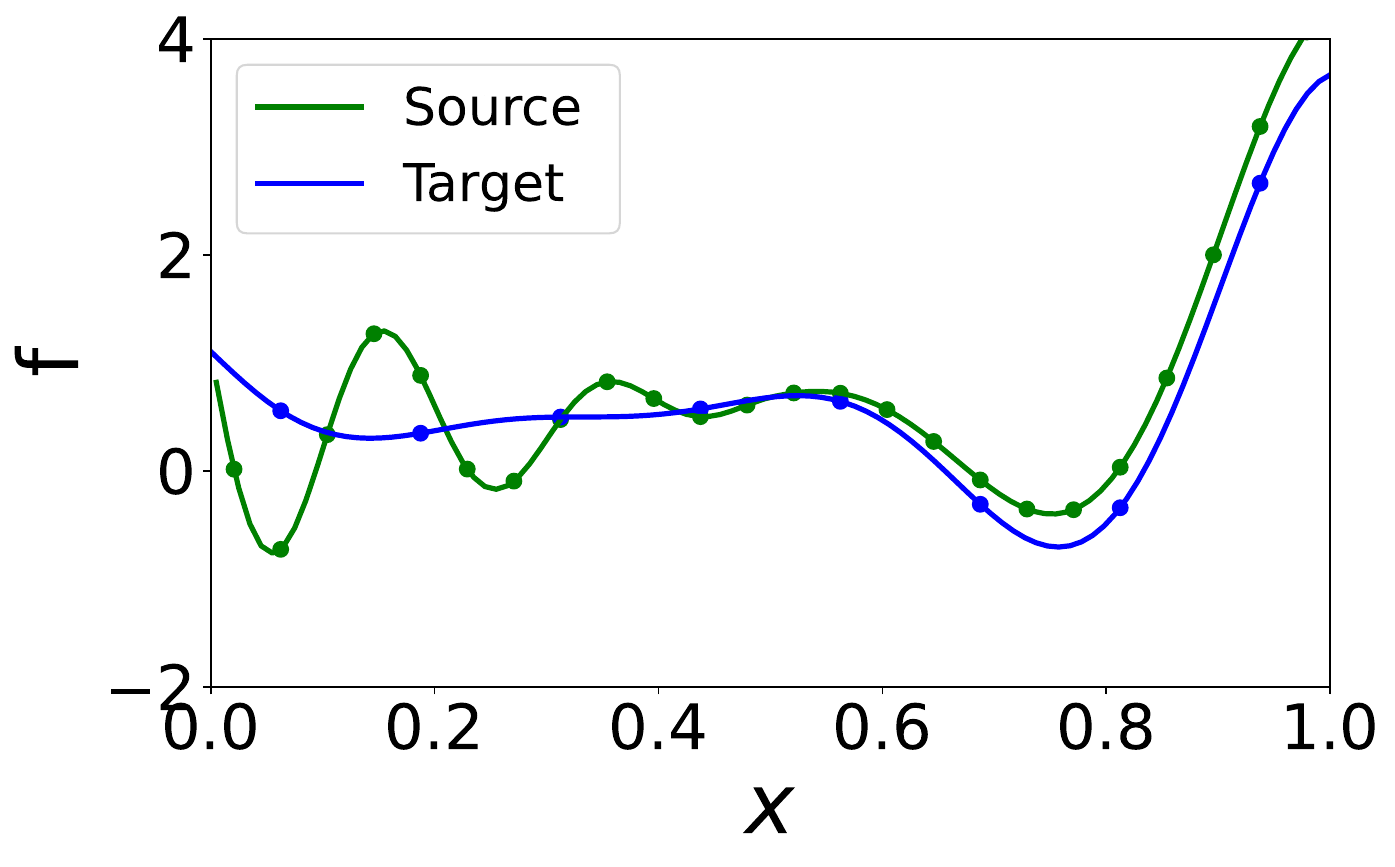}
		\caption{Visualizing the source and target functions (with corresponding design points) for our 1-d Forrester motivating example.}
		\label{fig: motivating example data}
	\end{figure}

	To explore such limitations, consider a motivating example using the one-dimensional Forrester function \cite{Forrester2008}. This function, which we treat as the target, is given by:
	\begin{equation}
		f_T({x}) = 0.2(6x-2)^2 \sin(12x - 4) + 0.5.
		\label{eq:fortarget}
	\end{equation}
	To mimic local transfer, where systems may be similar at certain parameters but considerably different at others, we adopt the source function as:
	\begin{equation}
		f_S({x}) = f_T({x}) + 0.6 \; \mathbb{I}_{\{x \geq 0.5\}}(x-0.5) + 1.6 \; \mathbb{I}_{\{x<0.5\}}(x-0.5)\sin(30x)\sqrt{5x-4} .
		\label{eq:forsource}
	\end{equation}
	\Figref{fig: motivating example data} visualizes the source and target functions. When $x \geq 0.5$, we see that the source function is nearly identical to the target; when $x < 0.5$, the source function becomes considerably different. This reflects the desired local transfer property expected in physical systems. To lessen the risk of negative transfer, a good model should identify the first region (where source and target are similar) as appropriate for transfer, and the latter region (where source and target are different) as detrimental for transfer. We collect training data on $n_S = 24$ equidistant design points for the source function, and data on $n_T = 8$ equidistant design points on the target (see \Figref{fig: motivating example data}).
	
	\begin{figure}[!t]
		\centering
		\includegraphics[width=0.96\linewidth]{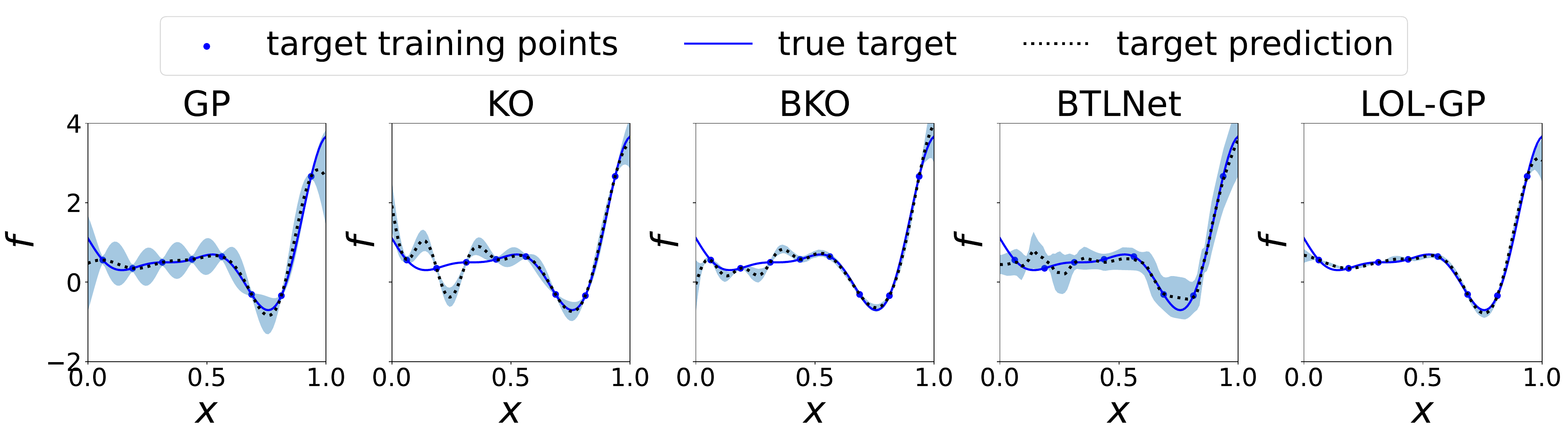}
		\caption{Visualizing the fitted surrogates (dotted: predictive mean, shaded: 95\% confidence region) with the true target function (solid line) for the standard GP and existing transfer learning surrogates.}
		\label{fig: motivating example prediction}
	\end{figure}
	
	\Figref{fig: motivating example prediction} shows the fitted benchmark surrogates: the standard GP, the KO \cite{KO2000} and BKO models \cite{BayesianKO2008}; the first model is trained using just target data, whereas the latter are trained using both source and target data. We see the standard GP yields poor predictions with high uncertainty, which is unsurprising due to its small sample size of $n_T =8$; some transfer would thus be beneficial here. The KO model \eqref{eq: KOtr}, which adopts {constant} transfer over the full domain, yields noticeably improved predictions over the region $[0.5, 1.0]$, where transfer is desirable. However, over $[0.0, 0.5]$, where the source and target are considerably different, the modeled transfer appears to hurt performance: the fitted surrogate yields poorer predictions with overly certain UQ compared to the standard GP. This highlights how the simple transfer model in KO may induce negative transfer by failing to restrict transfer over $[0.0, 0.5]$. The BKO model instead adopts a more flexible input-dependent transfer, where $\rho(x)$ follows a GP. From Figure \ref{fig: motivating example prediction}, BKO yields slight improvements over KO, but still has worse performance compared to the standard GP over $[0.0, 0.5]$. This suggests that, with limited target data, a highly flexible transfer model may also induce negative transfer. We further compare with a Bayesian transfer learning neural network benchmark called BTLNet (details on its architecture discussed later in Section \ref{sec: numerical experiments}). From Figure \ref{fig: motivating example prediction}, the predictions from BTLNet can be seen to be rather erratic throughout, perhaps due to potential overfitting of the transfer model given limited target data. 
	
	We present next the proposed LOL-GP, which aims to tackle such limitations. In particular, the LOL-GP makes use of a latent regularization model that learns where on the parameter space transfer may be beneficial and where it may be detrimental. In doing so, Figure \ref{fig: motivating example prediction} shows that it can provide considerably better performance over existing models.
	
	\section{The LOL-GP Model}
	\label{sec: model}
	
	We first present the LOL-GP for the general multi-source transfer learning set-up, where there are $L$ source systems and a single target system. We then investigate a natural extension of the LOL-GP for multi-fidelity surrogates.
	
	\subsection{Multi-source transfer}
	\label{subsec: multisystem}
	Consider first the general transfer learning set-up, where we have simulated data on $L \geq 1$ related source systems and wish to predict the target simulator. For the $l$-th source system, suppose its simulator (denoted as $f_{l}(\cdot)$) is run at inputs $\mathcal{X}_{l} = \{\bm{x}_{l,1}, \cdots, \bm{x}_{l,n_l}\} \subseteq \mathbb{R}^d$, yielding outputs $\bm{f}_l \in \mathbb{R}^{n_l}$. Further suppose the target simulator (denoted as $f_T(\cdot)$) is run at inputs $\mathcal{X}_T = \{\bm{x}_{1}, \cdots, \bm{x}_{n_T}\} \subseteq \mathbb{R}^d$, yielding outputs $\bm{f}_T \in \mathbb{R}^{n_T}$.
	
	The LOL-GP then adopts the following transfer learning model:
	\begin{align}
		\begin{split}
			&f_T(\bm{x}) = \sum_{l=1}^{L} \color{black}{\rho_l}(\bm{x}) \color{black} f_{l}(\bm{x}) + \delta (\bm{x}),\\
			&f_l(\bm{x}) \sim \mathcal{GP}\{0, k_l(\cdot,\cdot)\}, \quad \delta(\bm{x}) \sim \mathcal{GP}\{0, k_{\delta}(\cdot,\cdot)\},\\ 
			&{\color{black} \rho_l(\bm{x}) = \text{ReLU}\{\omega_l(\bm{x})\}}, \quad \omega_l(\bm{x}) \sim \mathcal{GP}\{0, k_{\omega_l}(\cdot,\cdot)\}, \quad l=1,\cdots,L.    
			\label{eq: multi-source formulation}
		\end{split}
	\end{align}
	Here, all GP priors are specified independently. We investigate next each line of the above specification. The first line models the target simulator as a weighted sum of the source simulators, where the transfer weights $\rho_l(\bm{x})$ may depend on input parameters $\bm{x}$. Here, $\delta(\bm{x})$ accounts for potential discrepancies between the target system and its weighted predictor using the $L$ source systems. The second line in \eqref{eq: multi-source formulation} places independent GP priors on the unknown response surfaces for source systems as well as the discrepancy term. The key novelty is the third line of \eqref{eq: multi-source formulation}, where the transfer function is fed through the rectified linear unit (ReLU) activation function $\text{ReLU}(\cdot)=\max(0,\cdot)$, which is widely used in neural network models \cite{agarap2018deep}. Here, the ReLU activation provides an appealing form for \textit{regularizing} the desired local transfer behavior. When the latent function $\omega_l(\bm{x})$ exceeds zero, transfer is permitted by setting the transfer weight as $\rho_l(\bm{x}) = \omega_l(\bm{x}) > 0$; when $\omega_l(\bm{x})$ falls below zero, transfer is restricted by zeroing out its transfer weight. Other common activation functions in the literature, e.g., the sigmoid or softplus functions \cite{han1995, dugas2000}, do not have such a ``zeroing-out'' effect, hence our use of the ReLU activation here.
	
	Further insights can be gleaned on the LOL-GP through the lens of the \textit{bias-variance trade-off} \cite{hastie2009elements}, which is broadly used in statistical learning. Consider the following general transfer learning framework in the one-source setting: $f_T(\mathbf{x}) = g\{f_S(\mathbf{x}), \mathbf{x}\} + \delta(\mathbf{x})$, where the first term $g\{f_S(\mathbf{x}),\mathbf{x}\}$ facilitates transfer from the source output $f_S(\mathbf{x})$ via the function $g$, and the second term is an additive discrepancy term accounting for lack-of-fit (following the KOH literature). In this sense, the KOH model \eqref{eq: KOtr} adopts a simple form for the general transfer function $g\{f_S(\mathbf{x}),\mathbf{x}\} = \rho f_S(\mathbf{x})$, which may make it \textit{biased} when the underlying transfer is more complex. The BKO model \eqref{eq: BKO} employs a more flexible form $g\{f_S(\mathbf{x}),\mathbf{x}\} = \rho(\mathbf{x}) f_S(\mathbf{x})$ where $\rho(\mathbf{x})$ follows a GP. Given limited target data and the presence of local transfer, however, such a transfer model may be overly flexible and may be prone to overfitting, i.e., high \textit{variance}. The LOL-GP \eqref{eq: multi-source formulation}, which uses the transfer form $g\{f_S(\mathbf{x}),\mathbf{x}\} = \text{ReLU}\{\omega(\mathbf{x})\}f_S(\mathbf{x})$, offers a compromise in this bias-variance trade-off, by regularizing the transfer model via the ReLU activation function. Such regularization is guided by prior knowledge of local transfer and is learned from data; using a GP prior on the latent function $\omega(\mathbf{x})$, its posterior distribution can identify where transfer may or may not be beneficial over the parameter space. As we see later, this can help temper the risk of negative transfer, allowing for \textit{robust} transfer learning in surrogate modeling.

The proposed transfer form in \eqref{eq: multi-source formulation} also has connections to multiplicative discrepancy modeling (see, e.g., \cite{kleiber2014model}). Here, we use this multiplicative form as an intuitive yet flexible model for transferring information from source to target. Such a form has two key advantages for the LOL-GP. First, it provides a natural mechanism for modeling local transfer, as transfer is restricted when the latent function $\omega_l(\mathbf{x})$ (which is fitted from data) does not exceed zero. Second, as shown next, this form facilitates efficient posterior predictive sampling via closed-form full conditional distributions. When the underlying local transfer behavior can be well-modeled by this multiplicative form, our model is expected to work well; we find this to be the case in later experiments and applications.

For predictions on the target system, one needs to investigate the posterior distribution of $f_T(\bm{x}_{\rm new})$, conditional on data from source and target systems, namely, $\bm{f}_1, \cdots, \bm{f}_L$ and $\bm{f}_T$, respectively. We will denote this as $[f_T(\bm{x}_{\rm new}) | \bm{f}_1, \cdots, \bm{f}_L,\bm{f}_T]$, where $[X]$ denotes the distribution of a random variable $X$. Unfortunately, unlike the standard GP equations \eqref{eq: gp prediction}, this desired posterior distribution cannot be evaluated in closed form. The LOL-GP does, however, admit nice structure for efficient Gibbs sampling \cite{geman1984stochastic}, a popular Markov-Chain Monte Carlo (MCMC) technique that cyclically updates model parameters via full conditional sampling. We show next that, with a careful selection of latent parameters $\boldsymbol{\Theta}$, such full conditional distributions can be derived in closed form, enabling efficient posterior sampling of $f_T(\bm{x}_{\rm new})$.

Assume for now that the GP model parameters (i.e., its kernel hyperparameters) are fixed; we discuss the estimation of such parameters later in Section \ref{sec:map}. Define the set of latent parameters $\boldsymbol{\Theta}$ as $\boldsymbol{\Theta} = \{\omega_1(\mathcal{X}_T), \cdots, \omega_L(\mathcal{X}_T), f_1(\mathcal{X}_T), \cdots, f_L(\mathcal{X}_T)\}$, where ${f}_l(\mathcal{X}) = [f_l(\bm{x})]_{\bm{x} \in \mathcal{X}}$ is the vector of outputs for the $l$-th source system at design points $\mathcal{X}$. For notational brevity, we presume that no design points are shared between source and target systems in the following derivation; Supplementary Materials SM1 provides analogous derivations for the setting with potentially shared design points.

Given $\text{data} = \{\bm{f}_1, \cdots, \bm{f}_L,\bm{f}_T\}$, we can derive the following full conditional distributions\footnote{In what follows, we denote $[\theta|\boldsymbol{\Theta}_-,\text{data}]$ as the full conditional distribution of parameter $\theta \in \boldsymbol{\Theta}$, conditioned on all remaining parameters on $\boldsymbol{\Theta}$ as well as training data.} on the latent weights in $\boldsymbol{\Theta}$:
\begin{align}
	\begin{split}
		[\omega_l(\bm{x}_i) | \boldsymbol{\Theta}_-,\text{data}] &\sim \pi_{i,l}\;\mathcal{N}_{\mathbb{R}^+}({\mu}_{i,l,+}, {\sigma}^2_{i,l,+})\\
		& \quad \quad  + (1-\pi_{i,l}) \;\mathcal{N}_{\mathbb{R}^-}({\mu}_{i,l,-}, {\sigma}^2_{i,l,-}), \quad i = 1, \cdots, n_T, \quad l = 1, \cdots, L,
		\label{eq:fullcond}
	\end{split}
\end{align}
Following notation from spike-and-slab priors \cite{ishwaran2005spike}, $\pi \Pi_1 + (1-\pi) \Pi_2$ denotes a two-mixture distribution with weight $\pi$ on distribution $\Pi_1$ and weight $1-\pi$ on distribution $\Pi_2$. Here, $\mathcal{N}_{\mathbb{R}^+}({\mu}, {\sigma}^2)$ denotes the normal distribution $\mathcal{N}(\mu,\sigma^2)$ truncated on the non-negative reals $\mathbb{R}^+$, and $\mathcal{N}_{\mathbb{R}^-}({\mu}, {\sigma}^2)$ denotes the same distribution truncated on the non-positive reals $\mathbb{R}^-$.

The full conditional weights $\{\pi_{i,l}\}_{i,l}$ in \eqref{eq:fullcond} can then be evaluated as:
\begin{equation}
	\pi_{i,l} = \Phi\left(\frac{\mu_{i,l, +}}{\sigma_{i,l, +}}\right) \Big/ \left[\Phi\left(\frac{-\mu_{i,l, -}}{\sigma_{i,l, -}}\right) + \Phi\left(\frac{\mu_{i,l, +}}{\sigma_{i,l, +}}\right) \right],
	\label{eq:fullcond weight}
\end{equation}
where the means $\{\mu_{i,l,+},\mu_{i,l,-}\}_{i,l}$ and variances $\{\sigma^2_{i,l,+},\sigma^2_{i,l,-}\}_{i,l}$ can be computed as:
\small
\begin{align}
	\begin{split}
		\mu_{i,l,-} &= \bm{k}_{\omega_l}^T(\mathcal{X}_{T}^{[-i]}, \bm{x}_{i}) \bm{K}_{\omega_l}(\mathcal{X}_{T}^{[-i]},\mathcal{X}_{T}^{[-i]})^{-1}\bm{\omega}_{[-i],l},\\
		\sigma^2_{i,l, -}&= k_{\omega_l}(\bm{x}_{i}, \bm{x}_i) - \bm{k}_{\omega_l}^T(\mathcal{X}_{T}^{[-i]}, \bm{x}_{i}) \bm{K}_{\omega_l}(\mathcal{X}_{T}^{[-i]},\mathcal{X}_{T}^{[-i]})^{-1}\bm{k}_{\omega_l}(\mathcal{X}_{T}^{[-i]}, \bm{x}_{i}),\\
		\mu_{i,l,+} &= \sigma^2_{i,l, +} 
		\left(\frac{\mu_{i,l, -}}{\sigma^2_{i,l,-}} + \frac{{e}_{i,l} - \bm{k}_{\delta}^T(\mathcal{X}_{T}^{[-i]}, \bm{x}_{i}) \bm{K}_{\delta}(\mathcal{X}_{T}^{[-i]}, \mathcal{X}_{T}^{[-i]})^{-1}\left(\bm{f}_{[-i], T} - \sum_{l} \bm{\rho}_{l}^{[-i]} \odot \bm{f}_{l}^{[-i]} \right)}{\left[k_{\delta}(\bm{x}_{i}, \bm{x}_i) - \bm{k}_{\delta}^T(\mathcal{X}_{T}^{[-i]}, \bm{x}_{i}) \bm{K}_{\delta}(\mathcal{X}_{T}^{[-i]}, \mathcal{X}_{T}^{[-i]})^{-1} \bm{k}_{\delta}(\mathcal{X}_{T}^{[-i]}, \bm{x}_{i})\right] / f_l(\bm{x}_{i})} \right),\\
		\sigma^2_{i,l,+} &= \left(\frac{1}{\sigma^2_{i,l, -}} + \frac{[f_l(\bm{x}_{i})]^2}{k_{\delta}(\bm{x}_{i}, \bm{x}_i) - \bm{k}_{\delta}^T(\mathcal{X}_{T}^{[-i]}, \bm{x}_{i}) \bm{K}_{\delta}(\mathcal{X}_{T}^{[-i]}, \mathcal{X}_{T}^{[-i]})^{-1} \bm{k}_{\delta}(\mathcal{X}_{T}^{[-i]}, \bm{x}_{i})}\right)^{-1}.
		\label{eq:condclosed}
	\end{split}
\end{align}
\normalsize
Here, $\mathcal{X}_{T}^{[-i]} = \mathcal{X}_T \setminus \bm{x}_i$ denotes the set of target design points without the $i$-th point, and $\bm{\omega}_{l}^{[-i]}$, $\bm{\rho}_{l}^{[-i]}$ and $\bm{f}_{l}^{[-i]}$ are the vectors for $\omega_l(\cdot)$, $\rho_l(\cdot)$ and $f_l(\cdot)$, respectively, at points $\mathcal{X}_T^{[-i]}$. Finally, the scalar $e_{i,l}$ is defined as $ e_{i,l} = {f}_{T}(\bm{x}_i) - \sum_{l'\not=l}{\rho}_{l'}(\bm{x}_i) {f}_{l'}(\bm{x}_i)$. Supplementary Materials SM2 provides a detailed derivation of the full conditional distribution \eqref{eq:fullcond}.

Similarly, we can derive the following full conditional distributions on the latent function values in $\boldsymbol{\Theta}$:
\begin{equation}
	[f_l(\bm{x}_i) | \boldsymbol{\Theta}_-,\text{data}] \sim \mathcal{N}(\mu_{i,l},{\sigma}^2_{i,l}), \quad i = 1, \cdots, n_T, \quad l = 1, \cdots, L.
	\label{eq:fullcond2}
\end{equation}
The posterior means $\{\mu_{i,l}\}_{i,l}$ and variances $\{\sigma^2_{i,l}\}_{i,l}$ take the closed form:
\begin{equation}
	{\mu}_{i,l} = \bm{k}_l^T(\mathcal{X}_{l \cup T}, \bm{x}_i) \bm{K}_{l \cup T}^{-1}
	\bm{f}_{l \cup T}, \quad {\sigma}^2_{i,l} = {k}_l(\bm{x}_i, \bm{x}_i) - \bm{k}_l^T(\mathcal{X}_{l \cup T}, \bm{x}_i)  \bm{K}_{l \cup T}^{-1} \bm{k}_l(\mathcal{X}_{l \cup T}, \bm{x}_i),
	\label{eq:condclosed2}
\end{equation}
where $\mathcal{X}_{l \cup T} = \mathcal{X}_{l} \cup \mathcal{X}_{T}$, $\bm{f}_{l \cup T} = [\bm{f}_l, \bm{f}_T]$, $\bm{k}_l(\bm{x}_i, \mathcal{X}_{l \cup T})= [ \bm{k}_l(\bm{x}_i, \mathcal{X}_l), \rho_l(\bm{x_i}) \bm{k}_l(\bm{x}_i, \mathcal{X}_T)]$, and:
\begin{align*}
	\bm{K}_{l \cup T} = 
	\begin{bmatrix} \bm{K}_l(\mathcal{X}_l, \mathcal{X}_l) & \left[\bm{1}_{n_l}\bm{\rho}_l^T(\mathcal{X}_T)\right] \odot \bm{K}_l(\mathcal{X}_l, \mathcal{X}_T) \\ \left[\bm{\rho}_l(\mathcal{X}_T)\bm{1}_{n_l}^T\right] \odot \bm{K}_l(\mathcal{X}_T, \mathcal{X}_l) &  \sum_{l^{\prime}=1}^{L} \bm{K}_{l^{\prime}}(\mathcal{X}_T, \mathcal{X}_T)+\bm{K}_{\delta}(\mathcal{X}_T, \mathcal{X}_T) \end{bmatrix}
\end{align*}
is the covariance matrix for $\bm{f}_{l \cup T}$. Here, $\odot$ denotes the Hadamard (entry-wise) product.

While the above looks quite involved, the key appeal is that the closed-form full conditional distributions \eqref{eq:fullcond} and \eqref{eq:fullcond2} can be sampled quickly, thus permitting \textit{efficient} Gibbs sampling of the posterior distribution $[\boldsymbol{\Theta}|\text{data}]$. Such a sampler proceeds as follows. First, the latent parameters in $\boldsymbol{\Theta}$ are initialized in the sample chain. Next, one sequentially samples (and subsequently updates) each parameter $\theta \in \boldsymbol{\Theta}$ from the full conditional distributions \eqref{eq:fullcond} and \eqref{eq:fullcond2}. Each full conditional step can be sampled quickly here, as it involves sampling either a normal or a two-mixture normal distribution; further analysis of computational complexity is provided in Section \ref{sec:comp}. These steps are then repeated for $B \gg 1$ iterations until the sample chain on $\boldsymbol{\Theta}$ converges to a stationary distribution. One can show \cite{gelman1995bayesian} that such a stationary distribution is indeed the posterior distribution $[\boldsymbol{\Theta}|\text{data}]$. Algorithm \ref{algorithm:GibbsSamplingPlusPrediciton} summarizes these steps for our Gibbs sampler.

Finally, with the above MCMC samples (denoted $\{\boldsymbol{\Theta}_{[b]}\}_{b=1}^B$) in hand, we can then sample from the desired posterior predictive distribution $[f_T(\bm{x}_{\rm new})|\text{data}]$. Note that:
\begin{equation}
	[f_T(\bm{x}_{\rm new})|\text{data}] = \int [f_T(\bm{x}_{\rm new})|\boldsymbol{\Theta},\text{data}] [\boldsymbol{\Theta}|\text{data}] d\boldsymbol{\Theta} =: \int \circled{A} \; \circled{B} \; d\boldsymbol{\Theta}.
	\label{eq:postpred}
\end{equation}
Here, the distribution $\circled{A}$ can be sampled via Monte Carlo as follows. First (i), draw a sample from $[\{f_l(\bm{x}_{\rm new})\}_{l=1}^L | \boldsymbol{\Theta},\text{data}]$, where each entry $[f_l(\bm{x}_{\rm new})|\boldsymbol{\Theta},\text{data}]$ follows independent normal distributions from the standard GP equations \eqref{eq: gp prediction}, conditional on $f_l(\mathcal{X}_{l \cup T})$ and using kernel $k_l$. Next (ii), draw a sample from $[\{\omega_l(\bm{x}_{\rm new})\}_{l=1}^L | \boldsymbol{\Theta},\text{data}]$, where $[\omega_l(\bm{x}_{\rm new})|\boldsymbol{\Theta},\text{data}]$ follows independent normal distributions from the standard GP equations \eqref{eq: gp prediction}, conditional on $\omega_l(\mathcal{X}_T)$ and using kernel $k_{\omega_l}$. Finally (iii), with $\boldsymbol{\Psi} := (\{f_l(\bm{x}_{\rm new})\}_{l=1}^L,\{\omega_l(\bm{x}_{\rm new})\}_{l=1}^L)$ in hand, sample $f_T(\bm{x}_{\rm new})$ from:
\small
\begin{align}
	\begin{split}
		&[f_T(\bm{x}_{\rm new}) | \boldsymbol{\Psi}, \boldsymbol{\Theta},\text{data}] \\
		& \sim \mathcal{N} \left\{ \sum_{l=1}^L \rho_l(\bm{x}_{\rm new}) {f}_l(\bm{x}_{\rm new}) + \bm{k}_{\delta}^T(\mathcal{X}_T, \bm{x}_{\rm new}) \bm{K}_{\delta}(\mathcal{X}_T, \mathcal{X}_T)^{-1}\left[ \bm{f}_T(\mathcal{X}_T) - \sum_{l=1}^L \bm{\rho}_l(\mathcal{X}_T)\odot\bm{f}_l(\mathcal{X}_T) \right], \right.\\
		&\qquad \qquad k_{\delta}(\bm{x}_{\rm new},\bm{x}_{\rm new}) - \bm{k}_{\delta}^T(\mathcal{X}_T, \bm{x}_{\rm new}) \bm{K}_{\delta}(\mathcal{X}_T, \mathcal{X}_T)^{-1} \bm{k}_{\delta}(\mathcal{X}_T, \bm{x}_{\rm new}) \Bigg\},
		\label{eq:postpred condi f_T}
	\end{split}
\end{align}
\normalsize
where $\rho_l(\bm{x}_{\rm new})={\rm ReLU}\{\omega(\bm{x}_{\rm new})\}$.
The distribution $\circled{B}$ has already been sampled via the MCMC chain $\{\boldsymbol{\Theta}_{[b]}\}_{b=1}^B$. Thus, from \eqref{eq:postpred}, the desired predictive distribution $[f_T(\bm{x}_{\rm new})|\text{data}]$ can be sampled by taking a sample $\boldsymbol{\Theta}_{[b]}$ from the MCMC chain $\{\boldsymbol{\Theta}_{[b]}\}_{b=1}^B$, then performing steps (i)-(iii) with $\boldsymbol{\Theta} =\boldsymbol{\Theta}_{[b]}$ to obtain a sample of $f_T(\bm{x}_{\rm new})$. Algorithm \ref{algorithm:GibbsSamplingPlusPrediciton} summarizes these steps for posterior predictive sampling.

It should be noted that there may be identifiability issues if one wishes to separately infer the transfer functions $\{\rho_l(\mathbf{x})\}_l$ and the discrepancy $\delta(\mathbf{x})$ from data, although such issues should not affect the target goal of prediction on $f_T$. This is akin to the KO model, which can be used for predictive purposes despite similar non-identifiability issues \cite{tuo2018prediction}. Care should be taken if one wishes to make inference on these functions separately; this should be discussed with practitioners to validate any learned similarities and/or discrepancies between systems.

\begin{algorithm}[!t]
	\caption{Multi-source LOL-GP: Gibbs sampling and posterior predictive sampling}
	\label{algorithm:GibbsSamplingPlusPrediciton}
	\begin{algorithmic}
		\STATE \textit{Require}: Source data $\{ (\mathcal{X}_l, \bm{f}_l)\}_{l=1}^{L}$, target data $(\mathcal{X}_T,\bm{f}_T)$, number of MCMC iterations $B$, initial MCMC parameters $\boldsymbol{\Theta}_{[0]}$.
		\STATE 
		\vspace{0.2cm}
		\underline{\textit{Gibbs Sampling of $[\boldsymbol{\Theta}|\textup{data}]$}}:
		\STATE $\bullet$ \; Initialize $\boldsymbol{\Theta} \leftarrow \boldsymbol{\Theta}_{[0]}$ in the MCMC chain.
		\FOR{$b = 1, \cdots, B$}
		\STATE $\bullet$ Set $\boldsymbol{\Theta} \leftarrow \boldsymbol{\Theta}_{[b-1]}$.
		
		\FOR{$l = 1, \cdots, L$}
		\FOR{$i = 1, \cdots, n_T$}
		\STATE $\bullet$ Update $\omega_l(\bm{x}_i)$ in $\boldsymbol{\Theta}$ by sampling from the full conditional distribution \eqref{eq:fullcond}.
		\STATE $\bullet$ Update $f_l(\bm{x}_i)$ in $\boldsymbol{\Theta}$ by sampling from the full conditional distribution \eqref{eq:fullcond2}.
		\ENDFOR
		\ENDFOR   
		\STATE $\bullet$ Update $\boldsymbol{\Theta}_{[b]} \leftarrow \boldsymbol{\Theta}$.
		\ENDFOR
		\STATE \textit{Return}: MCMC samples $\{\boldsymbol{\Theta}_{[b]}\}_{b=1}^B$.
		\vspace{0.1cm}
		\STATE \underline{\textit{Posterior Predictive Sampling of $[f_T(\bm{x}_{\rm new})|\textup{data}]$}}:
		\STATE $\bullet$ Denote $\{\omega_{l,[b]}\}_{b=1}^B$, $\{f_{l,[b]}\}_{b=1}^B$ and $\{f_{[b]}\}_{b=1}^B$ as the posterior samples for $\omega_l(\bm{x}_{\rm new})$, $f_l(\bm{x}_{\rm new})$ and $f_T(\bm{x}_{\rm new})$, respectively.
		\FOR{$b = 1, \cdots, B$}
		\FOR{$l = 1, \cdots, L$}
		\STATE $\bullet$ Sample 
		$\begin{aligned}[t]
			\omega_{l,[b]} \sim 
			& \; \mathcal{N}\left\{\bm{k}_{\omega_l}^T(\mathcal{X}_{T}, \bm{x}_{\text{new}}) \bm{K}_{\omega_l}(\mathcal{X}_{T}, \mathcal{X}_T)^{-1} \boldsymbol{\omega}_l(\mathcal{X}_{T}), \right. \\
			& \left. \quad \quad k_{\omega_l}(\bm{x}_{\text{new}}, \bm{x}_{\text{new}})- \bm{k}_{\omega_l}^T(\mathcal{X}_{T}, \bm{x}_{\text{new}}) \bm{K}_{\omega_l}(\mathcal{X}_{T}, \mathcal{X}_T)^{-1} \bm{k}_{\omega_l}(\mathcal{X}_{T}, \bm{x}_{\text{new}}) \right\}
		\end{aligned}$ \\
		with $\omega_l(\mathcal{X}_{T})$ taken from $\boldsymbol{\Theta}_{[b]}$.
		\STATE $\bullet$ Sample 
		$\begin{aligned}[t]
			f_{l,[b]} \sim 
			& \; \mathcal{N} \left\{\bm{k}_{l}^T(\mathcal{X}_{l \cup T}, \bm{x}_{\text{new}}) \bm{K}_{l}(\mathcal{X}_{l \cup T}, \mathcal{X}_{l \cup T})^{-1} f_l(\mathcal{X}_{l \cup T}), \right.\\
			& \left. \quad \quad k_{l}(\bm{x}_{\text{new}}, \bm{x}_{\text{new}}) - \bm{k}_{l}^T(\mathcal{X}_{l \cup T}, \bm{x}_{\text{new}}) \bm{K}_{l}(\mathcal{X}_{l \cup T}, \mathcal{X}_{l \cup T})^{-1} \bm{k}_{l}(\mathcal{X}_{l \cup T}, \bm{x}_{\text{new}}) \right\}
		\end{aligned}$\\
		with $f_l(\mathcal{X}_{T})$ from $\boldsymbol{\Theta}_{[b]}$.
		\ENDFOR
		\STATE $\bullet$ Sample $f_{[b]} $ from the full conditional distribution \eqref{eq:postpred condi f_T} with $\{f_{l,[b]}\}_l$ and $\{w_{l,[b]}\}_l$.
		\ENDFOR
		\STATE \textit{Return}: MCMC samples  $\{f_{[b]}\}_{b=1}^B$ on $f_T(\bm{x}_{\rm new})$.
	\end{algorithmic}
\end{algorithm}

\subsection{Multi-fidelity transfer} 
\label{subsec:multifid}
The LOL-GP can further be adapted for robust transfer learning in \textit{multi-fidelity} surrogate modeling. Here, the $L$ source systems $f_1(\cdot), \cdots, f_L(\cdot)$ are ranked in terms of \textit{increasing} fidelity levels, and the goal is to transfer learning from such lower-fidelity systems to predict a target high-fidelity system $f_T(\cdot)$.

A natural adaptation of the LOL-GP for this multi-fidelity setting is the model:
\begin{align}
	\begin{split}
		&f_{l+1}(\bm{x}) = \rho_l(\bm{x}) f_{l}(\bm{x}) + \delta_{l}(\bm{x}),\\
		&f_1(\bm{x}) \sim \mathcal{GP}\{0, k_1(\cdot,\cdot)\}, \quad \delta_l(\bm{x}) \sim \mathcal{GP}\{0, k_{\delta_l}(\cdot,\cdot)\},\\ 
		&{\color{black} \rho_l(\bm{x}) = \text{ReLU}\{\omega_l(\bm{x})\}}, \quad \omega_l(\bm{x}) \sim \mathcal{GP}\{0, k_{\omega_l}(\cdot,\cdot)\}, \quad l=1,\cdots,L.    
		\label{eq:LOLGP-MF}
	\end{split}
\end{align}
As before, $f_T(\cdot) = f_{L+1}(\cdot)$ is the highest-fidelity simulator, and all GP priors are specified independently. This can be viewed as an extension of the BKO model \eqref{eq: BKO} that accounts for potential local transfer from low to high fidelity levels. As with the earlier multi-source transfer model, transfer from the $l$-th to $(l+1)$-th fidelity level is permitted only when the latent function $\omega_l(\bm{x}) > 0$; the posterior distributions on $\{\omega_l(\cdot)\}_{l=1}^L$ can then identify beneficial regions for multi-fidelity transfer. As we see in later numerical experiments, when this local transfer behavior is present in multi-fidelity systems, the integration of such behavior can facilitate \textit{robust} multi-fidelity modeling that lessens the risk of negative transfer. 

Following Section \ref{subsec: multisystem}, we can derive an efficient Gibbs sampler on $[\boldsymbol{\Theta}|\text{data}]$ for this multi-fidelity transfer model. Let $\mathcal{X}_l$ and $\bm{f}_l=f_l(\mathcal{X}_l)$ be the design points and simulated outputs for the $l$-th fidelity system, respectively. Assume for now that the GP model parameters (i.e., its kernel hyperparameters) are fixed; its estimation is discussed later in Section \ref{sec:map}. Let $\mathcal{X}_{(l)} = \cup_{l^{\prime}=l}^{L+1}\mathcal{X}_{l^{\prime}}$ denote the augmented design consisting of all design points with fidelity level $l$ or higher. With this, redefine the set of latent parameters $\boldsymbol{\Theta}$ as $\boldsymbol{\Theta} = \{\boldsymbol{\Theta}_1, \cdots, \boldsymbol{\Theta}_L\}$, where $\boldsymbol{\Theta}_l = \{\boldsymbol{\omega}_l(\mathcal{X}_{(l+1)}), \bm{f}_l(\mathcal{X}_{(l+1)})\}$. Using a similar approach as the multi-source setting, we can derive closed-from full conditional distributions on the latent parameters $\boldsymbol{\Theta}$, which can be used for efficient Gibbs sampling on the posterior distribution $[\boldsymbol{\Theta| \text{data}}]$. For brevity, this Gibbs sampler (with derivations of its full conditional distributions) are provided in Supplementary Materials SM3 (see Algorithm SM3.1).

With this Gibbs sampler for $[\boldsymbol{\Theta}|\text{data}]$, we can then \textit{sequentially} sample the posterior distribution $[f_{l}(\bm{x}_{\rm new})| \text{data}]$, from lowest fidelity (i.e., $l=1$) to highest fidelity (i.e., $l=L+1$). This is achieved by first rewriting the desired posterior distribution $[f_{L+1}(\bm{x}_{\rm new})| \text{data}]$ as:
\begin{align}
	\begin{split}
		[f_{L+1}(\bm{x}_{\rm new})|\text{data}] 
		&= \int [f_{L+1}(\bm{x}_{\rm new})|f_{L}(\bm{x}_{\rm new}),\boldsymbol{\Theta}, \text{data}] \\
		& \qquad \qquad [f_{L}(\bm{x}_{\rm new})|\bm{\Theta}, \text{data}] [\boldsymbol{\Theta}|\text{data}] \; df_{L}(\bm{x}_{\rm new}) \; d\boldsymbol{\Theta}.
	\end{split}
\end{align}
Note that the distribution $[f_{L}(\bm{x}_{\rm new})|\bm{\Theta}, \text{data}]$ above can be recursively decomposed via the following equation:
\begin{align}
	\begin{split}
		[f_{l+1}(\bm{x}_{\rm new})|\bm{\Theta}, \text{data}] &= \int [f_{l+1}(\bm{x}_{\rm new})|f_{l}(\bm{x}_{\rm new}),\boldsymbol{\Theta}, \text{data}] [f_{l}(\bm{x}_{\rm new})|\bm{\Theta}, \text{data}] \; df_{l}(\bm{x}_{\rm new})\\
		& =: \int \circled{A}_{l+1} \; [f_{l}(\bm{x}_{\rm new})|\bm{\Theta}, \text{data}] \; df_{l}(\bm{x}_{\rm new}),
		\quad l = 1, \cdots, L-1.
		\label{eq:recurse for pred multi-fidelity}
	\end{split}
\end{align}
Here, $\circled{A}_{l+1}$ denotes the distribution $[f_{l+1}(\bm{x}_{\rm new})|f_{l}(\bm{x}_{\rm new}),\boldsymbol{\Theta}, \text{data}]$. Applying the identity \eqref{eq:recurse for pred multi-fidelity} recursively for $l = L-1, L-2, \cdots, 1$, one can write the posterior $[f_{L+1}(\bm{x}_{\rm new})| \text{data}]$ as:
\begin{equation}
	[f_{L+1}(\bm{x}_{\rm new})| \text{data}] = \int \circled{A}_{L+1} \cdots \circled{A}_1 \circled{B} \; df_{L}(\bm{x}_{\rm new}) \cdots df_{1}(\bm{x}_{\rm new}) \; d\boldsymbol{\Theta}, 
	\label{eq:cond postpred multi-fidelity}
\end{equation}
where $\circled{A}_1 := [f_1(\bm{x}_{\rm new})|\boldsymbol{\Theta}, \text{data}]$ and $\circled{B} := [\boldsymbol{\Theta}|\text{data}]$ denotes the posterior distribution of latent parameters.

Equation \eqref{eq:cond postpred multi-fidelity} provides the roadmap for efficient posterior sampling from the desired predictive distribution $[f_{L+1}(\bm{x}_{\rm new})|\text{data}]$. One can show that the conditional distributions $\{\circled{A}_l\}_{l=1}^{L+1}$ can be efficiently sampled via a straight-forward Monte Carlo algorithm (details provided in Supplementary Materials SM3). The distribution $\circled{B}$ can be sampled via the earlier MCMC approach. With this, Algorithm SM3.1 in Supplementary Materials SM3 outlines the full posterior predictive sampling algorithm for this multi-fidelity set-up.

\section{Methodological Developments}
\label{sec: fitting and prediction}
Next, we outline methodological developments for practical implementation of the LOL-GP. We first present an efficient framework for optimizing kernel hyperparameters, then discuss our model's computational complexity and how this can be sped up in the multi-fidelity setting via a nested experimental design.

\subsection{Optimization of kernel hyperparameters}
\label{sec:map}
In the previous section, we presumed that the LOL-GP kernel hyperparameters (i.e., hyperparameters for kernels $k_{\delta}$, $\{k_l\}_{l=1}^L$ and $\{k_{\omega_l}\}_{l=1}^L$ in the multi-source case, and hyperparameters for kernels $\{k_{\delta_l}\}_{l=1}^L$, $k_1$ and $\{k_{\omega_l}\}_{l=1}^L$ in the multi-fidelity case) are known when deriving the sampler for posterior prediction. Such hyperparameters are, however, unknown in practice, and need to be optimized from data. Given the potentially large number of hyperparameters, careful consideration is needed on how these hyperparameters can be optimized efficiently. {Here, a fully Bayesian approach can be costly, as such hyperparameters do not admit closed-form full conditional distributions and may require long sample chains using off-the-shelf MCMC samplers for sufficient mixing. This can be undesirable for applications where efficient emulators are needed.} Further, a standard maximum likelihood or maximum-a-posteriori (MAP) approach may also be challenging, as both the likelihood and the marginal likelihood do not admit a closed form due to the presence of the latent weight functions $\{\omega_l(\cdot)\}_{l=1}^L$. One thus requires numerical approximations of the objective function for maximization, which can again be costly. For efficient model training, we elect for an approximate MAP procedure (described below) using the inferred latent parameters $\boldsymbol{\Theta}$ from Section \ref{sec: model}.

Let $\boldsymbol{\Xi} = \{\boldsymbol{\Xi}_{\delta},\boldsymbol{\Xi}_f,\boldsymbol{\Xi}_{\omega}\}$ denote the set of kernel hyperparameters to optimize. For the multi-source case, $\boldsymbol{\Xi}_{\delta}$, $\boldsymbol{\Xi}_{f}$ and $\boldsymbol{\Xi}_{\omega}$ consist of hyperparameters for kernels $k_\delta$, $\{k_l\}_{l=1}^L$ and $\{k_{\omega_l}\}_{l=1}^L$, respectively; for the multi-fidelity case, these consist of hyperparameters for kernels $\{k_{\delta_l}\}_{l=1}^L$, $k_1$ and $\{k_{\omega_l}\}_{l=1}^L$, respectively. Let $[\bm{\Xi}_{\delta}]$, $[\bm{\Xi}_f]$ and $[\bm{\Xi}_{\omega}]$ denote the respective prior distributions on parameters $\boldsymbol{\Xi}$. With this, we employ the optimization formulation:
\begin{align}
	\label{eq: log post}
	\hat{\bm{\Xi}} := \arg\max_{\bm{\Xi}}  \log [\bm{\Xi}|\text{data}, \hat{\boldsymbol{\Theta}}]
	& = \arg\max_{\bm{\Xi}}  \log\left\{ [\text{data} | \hat{\boldsymbol{\Theta}}, \boldsymbol{\Xi}] \; [\hat{\boldsymbol{\Theta}}| \boldsymbol{\Xi}] \; [\boldsymbol{\Xi}] \right\}.
\end{align}
Here, $\hat{\bm{\Theta}}$ is a point estimate of the latent parameters $\boldsymbol{\Theta}$, which we take as the posterior mean of the MCMC samples $\{\boldsymbol{\Theta}_{[b]}\}_{b=1}^B$ from Section \ref{sec: model}.  $\hat{\bm{\Xi}}$ can thus be viewed as an approximate MAP estimator of the kernel parameters $\bm{\Xi}$, using the plug-in estimator $\hat{\bm{\Theta}}$ for $\bm{\Theta}$.

The key benefit of using this estimator is that, under the LOL-GP, the terms $[\text{data} | \bm{\Xi}, \hat{\boldsymbol{\Theta}}]$ and $[\hat{\boldsymbol{\Theta}}|\bm{\Xi}]$ in \eqref{eq: log post} both admit analytic \textit{closed-form} expressions. In particular, such terms reduce to Gaussian likelihood expressions in both the multi-source and multi-fidelity cases; their full expressions are provided in Supplementary Materials SM4.  With this, the objective function in \eqref{eq: log post} can be evaluated analytically, facilitating efficient optimization via state-of-the-art nonlinear optimization algorithms. Such a closed-form objective further permits the use of \textit{automatic differentiation} \cite{baydin2018automatic}, which allows for quick computation of objective gradients without the need for deriving gradients by hand. In our later implementation, we made use of the L-BFGS algorithm \cite{nocedal1999numerical} with automatic differentiation to optimize \eqref{eq: log post}. To contrast, a standard maximum likelihood or MAP estimation of $\bm{\Xi}$ can be more costly due to the need for numerical approximation of the objective. {This efficiency also permits random restarts of the optimization procedure, which is important for overcoming local optima in hyperparameter optimization. Five random restarts are used in our later numerical experiments.}

Finally, we note that this hyperparameter optimization procedure does not aim to recover some underlying set of ``true'' hyperparameters. Rather, such optimization is geared towards improving the predictive performance of our model. This distinction is particularly important given the known identifiability issues in inferring covariance parameters for GP models \cite{zhang2004inconsistent}.

\subsection{Computational complexity}
\label{sec:comp}
Next, we investigate the computational complexity for fitting the LOL-GP, for both Gibbs sampling and hyperparameter optimization, We present this for first the multi-source setting then the multi-fidelity setting. Table \ref{tab: complexity} summarizes this complexity analysis.

Consider first the \textit{multi-source} transfer model from Section \ref{subsec: multisystem}. For Gibbs sampling, note that the matrix inverses in Algorithm \ref{algorithm:GibbsSamplingPlusPrediciton} can be pre-computed prior to sampling; this requires a total work of $\mathcal{O}\left\{n_T^3 + \sum_{l=1}^Ln_l^3 \right\}$ using the decremental algorithm in \cite{pang2023enhanced}. This same cost is incurred within each objective evaluation of \eqref{eq: log post} for hyperparameter estimation. With this computed, each Gibbs sampling step is dominated by the matrix operations in evaluating $\{{\mu_{i,l,-}\}_{i=1}^{n_T}}_{l=1}^L$, $\{{\mu_{i,l,+}\}_{i=1}^{n_T}}_{l=1}^L$ and $\{{\mu_{i,l}\}_{i=1}^{n_T}}_{l=1}^L$ (see Equations \eqref{eq:condclosed} and \eqref{eq:condclosed2}); this amounts to a total complexity of $
\mathcal{O}\left\{3L n_T^2 + n_T \sum_{l=1}^L n_l \right\}$ per Gibbs step. Thus, the bottleneck in model training lies within the earlier computation of matrix inverses, which is unsurprising given similar cubic bottlenecks for standard GPs \cite{li2024trigonometric}. 

Consider next the \textit{multi-fidelity} transfer model in Section \ref{subsec:multifid}. For its Gibbs sampler (see Algorithm SM3.1 in Supplementary Materials SM3), the matrix inverses can similarly be pre-computed prior to Gibbs sampling; this incurs a total work of $\mathcal{O}\left\{(\sum_{l=1}^{L+1}n_{(l)})^3\right\}$ using the decremental algorithm in \cite{pang2023enhanced}, where $n_{(l)}$ is the size of $\mathcal{X}_{(l)}$. The same cost is required for each objective evaluation of \eqref{eq: log post} for hyperparameter estimation. With this computed, each Gibbs sampling step is dominated by the matrix operations in evaluating $\{{\mu_{i,l,-}\}_{i=1}^{n_l}}_{l=1}^{L}$, $\{{\mu_{i,l,+}\}_{i=1}^{n_l}}_{l=1}^{L}$ and $\{{\mu_{i,l}\}_{i=1}^{n_l}}_{l=1}^L$(see Equations SM3.2 and SM3.4); this incurs a total complexity of $\mathcal{O}\left\{ \sum_{l=1}^{L}2n_{(l+1)}^2+n_{(l+1)}n_{\rm aug}\right\}$ per Gibbs step, where $n_{\rm aug}=\sum_{l=1}^{L+1}n_{(l)}$. Again, not surprisingly, the bottleneck for model training lies within the earlier computation of matrix inverses. Compared to the multi-source setting, however, this cost of matrix inverses is considerably higher for the multi-fidelity setting; the former requires $\mathcal{O}\left\{n_T^3 + \sum_{l=1}^{L}n_l^3\right\}$ work, whereas the latter requires $\mathcal{O}\left\{(\sum_{l=1}^{L+1}n_l)^3\right\}$ work. Fitting the multi-fidelity LOL-GP can thus be costly when the sum of sample sizes over all fidelity levels grows large. We explore next a nested experimental design approach for reducing this cost in the multi-fidelity setting.

\begin{table}[!t]
	\renewcommand{\arraystretch}{2.0}
	\setlength\tabcolsep{5pt}
	\setlength\abovecaptionskip{0cm}  
	\caption{Computational complexity analysis for training the multi-source and multi-fidelity LOL-GP. Provided are the costs (in big-$\mathcal{O}$) for each Gibbs sampling iteration (Sections \ref{subsec: multisystem} and \ref{subsec:multifid}) and each objective evaluation for hyperparameter optimization (Section \ref{sec:map}).}
	\label{tab: complexity}
	\centering
	\begin{tabular}{c cccc}
		\toprule
		& Non-nested design & Nested design \\
		\toprule
		Multi-source (Gibbs) & $\mathcal{O}\left\{3Ln_T^2 + n_T \sum_{l=1}^L n_l\right\}$ & $\mathcal{O}\left\{2Ln_T^2\right\}$\vspace{1em}\\ 
		{\makecell[c]{ Multi-source\\
				(Hyperparameter optimization)}} & $\mathcal{O}\left\{n_T^3 + \sum_{l=1}^{L}n_l^3\right\}$ & $\mathcal{O}\left\{n_T^3 + \sum_{l=1}^{L}n_l^3\right\}$ \\ [1.0ex]
		\hline
		Multi-fidelity (Gibbs) & $\mathcal{O}\left\{ \sum_{l=1}^{L}2n_{(l+1)}^2+n_{(l+1)}n_{\rm aug} \right\}$ & $\mathcal{O}\left\{\sum_{l=1}^L 2n_{l+1}^2\right\}$ \vspace{1em}\\
		{\makecell[c]{ Multi-fidelity\\
				(Hyperparameter optimization)}} & $\mathcal{O}\left\{\left(\sum_{l=1}^{L+1}n_l\right)^3\right\}$ & $\mathcal{O}\left\{\sum_{l=1}^{L+1}n_l^3\right\}$ \\ [1.0ex]
		\toprule
	\end{tabular}
\end{table}

\subsection{Nested experimental design}
\label{sec:nested}
We present next a nested experimental design approach for the multi-fidelity LOL-GP to reduce the aforementioned $\mathcal{O}\left\{(\sum_{l=1}^{L+1}n_l)^3\right\}$ cost for model training. This builds off of related work on recursive multi-fidelity modeling via nested designs (see, e.g., \cite{RecursiveMF2014,ji2023graphical}), which can be extended for our local transfer model. In what follows, we explore nested designs for only the multi-fidelity setting as its training cost is more burdensome; similar nested designs can be adapted for the multi-source setting, but this leads to only minor computational speed-ups (see Supplementary Materials SM5). 

We first introduce the notion of nested designs for the multi-fidelity LOL-GP. The lower-fidelity points $\mathcal{X}_1, \cdots, \mathcal{X}_L$ and the high-fidelity points $\mathcal{X}_T = \mathcal{X}_{L+1}$ are deemed \textit{nested} if $\mathcal{X}_{L+1} \subseteq \mathcal{X}_{L} \subseteq \cdots \subseteq \mathcal{X}_{1}$, i.e., higher-fidelity points are nested within lower-fidelity ones. Given this nested structure, consider the following recursive reformulation of the multi-fidelity LOL-GP:
\begin{align}
		f_{l+1}(\bm{x}) = \rho_{l} (\bm{x}) \tilde{f}_{l}(\bm{x}) + \delta_{l} (\bm{x}), \quad \tilde{f}_{l}(\bm{x}) \sim \mathcal{N}\{\tilde{\mu}_{l}(\bm{x}), \tilde{\sigma}_{l}^2(\bm{x})\}, \quad l = 1, \cdots, L,
		\label{eq:recurse formulation multi-fidelity}
\end{align}
where $\rho_l(\bm{x})$ and $\delta_l(\bm{x})$ follow the same priors from Equation \eqref{eq:LOLGP-MF}. Here, the mean and variance terms $\tilde{\mu}_{l}(\bm{x})$ and $\tilde{\sigma}_{l}^2(\bm{x})$ are defined recursively as:
\small
\begin{align}
	\begin{split}
		&\tilde{\mu}_{l}(\bm{x}) = \rho_{l-1}(\bm{x})\tilde{\mu}_{l-1}(\bm{x}) + \bm{k}_l^T(\mathcal{X}_l, \bm{x}) \bm{K}_l^{-1}(\mathcal{X}_l, \mathcal{X}_l) [\bm{f}_l - \boldsymbol{\rho}_{l-1}(\mathcal{X}_l) \odot \bm{f}_{l-1}(\mathcal{X}_l)],\\
		&\tilde{\sigma}_{l}^2(\bm{x}) = \rho_{l-1}^2(\bm{x}) \tilde{\sigma}^2_{l-1}(\bm{x}) + k_l(\bm{x}, \bm{x}) - \bm{k}_l^T(\mathcal{X}_l, \bm{x}) \bm{K}_l^{-1}(\mathcal{X}_l, \mathcal{X}_l) \bm{k}_l(\mathcal{X}_l, \bm{x}), \quad l = 1, \cdots, L.
		\label{eq: nested multi-fidelity prediction}
	\end{split}
\end{align}
\normalsize

The intuition behind this recursive formulation is as follows. Here, $\tilde{f}_{l}(\cdot)$ can be interpreted as the GP model for the $l$-th fidelity level, conditional on simulated data at that level. The recursion arises from \eqref{eq: nested multi-fidelity prediction}, where $\tilde{\mu}_l(\bm{x})$ and $\tilde{\sigma}^2_l(\bm{x})$ are defined in terms of their respective functions at the $(l-1)$-th fidelity level. One can show (via an analogous argument from \cite{RecursiveMF2014}) that, with nested design points, the posterior predictive distribution $[f_T(\bm{x}_{\rm new})|\text{data}]$ from the multi-fidelity LOL-GP \eqref{eq:LOLGP-MF} yields the same posterior predictive distribution from the recursive reformulation \eqref{eq:recurse formulation multi-fidelity}, {thus allowing for quicker computation.}

The recursive form \eqref{eq:recurse formulation multi-fidelity} with nested design points allows for two nice computational speed-ups. First, for optimizing kernel hyperparameters (Section \ref{sec:map}), it can be shown (see Supplementary Materials SM4) that the likelihood $[\text{data}|\boldsymbol{\Xi},\hat{\boldsymbol{\Theta}}]$ takes the form:
\begin{equation}
	[\text{data}|\boldsymbol{\Xi},\hat{\boldsymbol{\Theta}}] = \frac{1}{|\bm{C}_{1:(L+1)}|}\exp\left\{-\frac{1}{2}\bm{f}_{1:(L+1)}^T \bm{C}_{1:(L+1)}^{-1} \bm{f}_{1:(L+1)}\right\},
\end{equation}
where $\bm{f}_{1:(L+1)}$ is the data vector over all fidelity levels and $\bm{C}_{1:(L+1)}$ is its prior covariance matrix. Previously, each evaluation of this likelihood incurs $\mathcal{O}\left\{ (\sum_{l=1}^{L+1} n_l)^3 \right\}$ work, which forms the main bottleneck for hyperparameter optimization. Using the recursive formulation with nested designs, this likelihood simplifies to:
\begin{align}
	[\text{data} | \bm{\Xi}, \hat{\boldsymbol{\Theta}}] \propto 
	\prod_{l=1}^{L+1}\frac{1}{|\bm{K}_{l}(\mathcal{X}_l, \mathcal{X}_l)|}\exp\left\{-\frac{1}{2}\bm{z}_l^T \bm{K}_{l}^{-1}(\mathcal{X}_l, \mathcal{X}_l)\bm{z}_l\right\},
	\label{eq:likelihood of data for recursive multi-fidelity}
\end{align}
where $\bm{z}_1 = \bm{f}_1$ and $\bm{z}_l = \bm{f}_l - \boldsymbol{\rho}_{l-1}(\mathcal{X}_l) \odot \bm{f}_{l-1}(\mathcal{X}_l)$ for $l = 2, \cdots, L+1$. Equation \eqref{eq:likelihood of data for recursive multi-fidelity} thus factorizes the objective function over fidelity levels, which allows for separate optimization of hyperparameters at \textit{each} fidelity level. The matrix inverses $\{\bm{K}_l(\mathcal{X}_l, \mathcal{X}_l)\}_{l=1}^L$ can further be computed separately, which reduces the objective evaluation cost for hyperparameter optimization from $\mathcal{O}\left\{(\sum_{l=1}^{L+1}n_l)^3\right\}$ to $\mathcal{O}\left\{ \sum_{l=1}^{L+1} n_l^3\right\}$. A similar reduction holds for the pre-computation of matrix inverses prior to Gibbs sampling. Second, within each Gibbs sampling step (see Algorithm SM3.1 in Supplementary Materials SM3), one can bypass the sampling step for $[f_l(\bm{x}_i)|\boldsymbol{\Theta}_{-}, \text{data}]$ (see Equation SM3.3), as this is known with certainty using nested design points. The complexity of each Gibbs sampling step then reduces to $\mathcal{O}\left\{\sum_{l=1}^L 2n_{l+1}^2\right\}$. Table \ref{tab: complexity} summarizes the above computational speed-ups using nested designs.

\section{Numerical Experiments}
\label{sec: numerical experiments}
We now explore the proposed models in a suite of numerical experiments. The following existing methods will be used as benchmarks for comparison:
\begin{itemize}[leftmargin=*]
	\item \textbf{GP}: This is the standard GP model, using \textit{only} data from the target system with no transfer learning. Here, the standard anisotropic squared-exponential kernel \cite{gramacy2020surrogates} is used, with kernel hyperparameters fitted via maximum likelihood.
	\item \textbf{KO}: This is the Kennedy-O'Hagan model \cite{KO2000} with a constant correlation parameter $\rho$; its multi-fidelity form follows \eqref{eq: KO}, and its multi-source form follows \eqref{eq: multi-source formulation} with $\rho_l(\bm{x}) = \rho$. All GPs employ anisotropic squared-exponential kernels \cite{gramacy2020surrogates}, with kernel hyperparameters and $\rho$ fitted via maximum likelihood.
	\item \textbf{BKO}: This is the Bayesian Kennedy-O'Hagan model \cite{BayesianKO2008} with $\rho(\bm{x})$ varying in $\bm{x}$; its multi-fidelity form follows \eqref{eq: BKO}, and its multi-source form follows \eqref{eq: multi-source formulation} with $\rho_l(\bm{x}) = \omega_l(\bm{x})$. All GPs employ anisotropic squared-exponential kernels \cite{gramacy2020surrogates}, with kernel hyperparameters fitted via maximum likelihood.
	\item \textbf{BTLNet}: {This is a Bayesian neural network baseline model for transfer learning, adapted from the fully-connected Bayesian neural network in \cite{meng2021multi} using standard ReLU activations. Such a model is fit using mean-field variational inference, following \cite{meng2021multi}. Supplementary Materials SM6 provides implementation details on its modeling architecture, as well as additional experimental results with other choices of architectures and activation functions.}
\end{itemize}
\noindent {Here, all methods requiring MCMC are run for 10,000 iterations (with 1,000 iterations discarded as burn-in), and convergence is checked via standard MCMC diagnostic metrics \cite{gelman1995bayesian}.}

To evaluate surrogate performance, we employ the following two metrics:
\begin{itemize}[leftmargin=*]
	\item Root-mean-squared-error (RMSE): The RMSE, defined as $\sqrt{{n_{\text{test}}^{-1}} \sum_{i=1}^{n_{\text{test}}} [\hat{f}_T(\tilde{\bm{x}}_i) - f_{T}(\tilde{\bm{x}}_i)]^2}$, measures \textit{point} prediction accuracy on the target function $f_T(\cdot)$. Here, $\tilde{\bm{x}}_1, \cdots, \tilde{\bm{x}}_{n_{\rm test}}$ are test points, and $\hat{f}_T(\tilde{\bm{x}})$ is the point prediction at $\tilde{\bm{x}}$.
	
	\item Continuous ranked probability score (CRPS; \cite{gneiting2007strictly}): The CRPS is defined as the following scoring rule:
	\begin{equation}
		{n_{\text{test}}^{-1}} \sum_{i=1}^{n_{\text{test}}} \int \left[ F_{Y_T(\tilde{\bm{x}}_i)}(u) - \mathbb{I}_{\{u \geq f_{T}(\tilde{\bm{x}}_i)\}} \right]^2 du.
	\end{equation}
	Here, $Y_T(\tilde{\bm{x}})$ is the \textit{probabilistic} (i.e., random) predictor at test point $\tilde{\bm{x}}$, with $F_{Y_T(\tilde{\bm{x}})}(u)$ its cumulative distribution function.
\end{itemize}
For both metrics, smaller values indicate better performance. A smaller RMSE suggests better point prediction accuracy, whereas a smaller CRPS suggests better probabilistic predictions.

\subsection{Multi-source transfer experiments}
\begin{figure}[!t]
	\centering
	\includegraphics[width=0.40\linewidth]{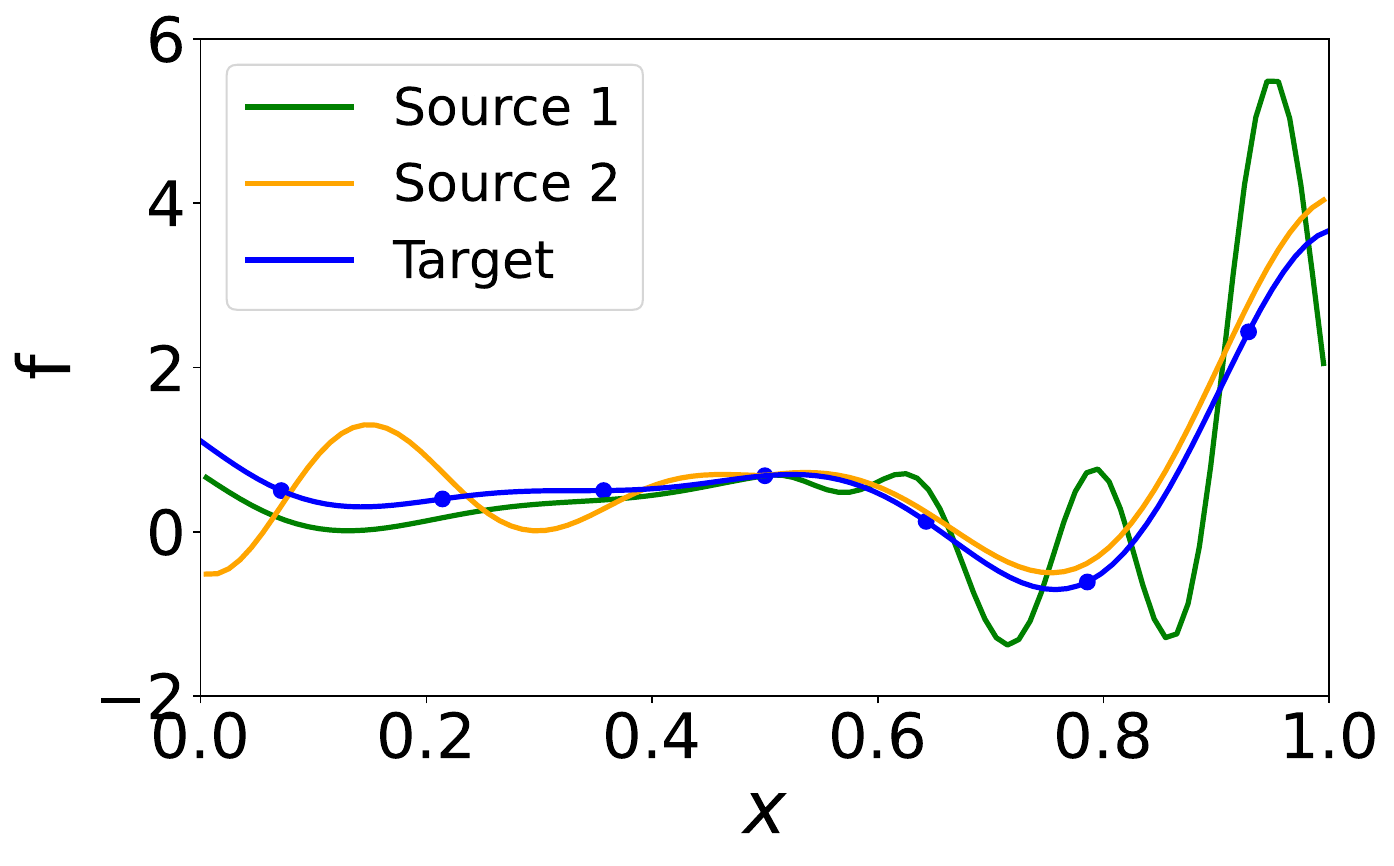}
	\caption{Visualizing the source and target functions for the 1-d multi-source Forrester experiment.}
	\label{fig: MS synthetic data}
\end{figure}

We first consider numerical experiments for multi-source transfer with two source systems ($f_1$ and $f_2$) and one target system ($f_T$). The first experiment, which builds off of the earlier 1-d Forrester function \cite{Forrester2008}, uses:
\begin{align}
	\begin{split}
		f_1(x) &= f_T(x) + 0.4\mathbb{I}_{\{x<0.5\}}(x-0.5) + 1.6\mathbb{I}_{\{x \geq 0.5\}}(x-0.5)\cos(40x)(5x-1), \\
		f_2(x) &= f_T(x) + 0.4\mathbb{I}_{\{x \geq 0.5\}}(x-0.5) + 1.6\mathbb{I}_{\{x<0.5\}}(x-0.5)\cos(20x)\sqrt{4-5x}, \\
		f_T(x) &= 0.2(6x-2)^2 \sin(12x - 4) + 0.5,
		\label{eq:forrester}
	\end{split}
\end{align}
where $f_T$ is the original Forrester function in \cite{Forrester2008}. Figure \ref{fig: MS synthetic data} visualizes these functions. Note that local transfer is implicitly captured here: the first source $f_1$ mimics the target function within the local region $x \in [0,0.5]$, whereas the second source $f_2$ mimics the target within $x \in [0.5,1.0]$. The second experiment, which builds off of the 5-d Friedman function \cite{Friedman1991}, uses:
\begin{align}
	\begin{split}
		f_1(\bm{x}) &= 2 \sin(\pi x_1 x_2) + 2.2 x_4 - x_5 \\
		& \qquad \qquad  + 8(x_3-0.5)^2[1.2\mathbb{I}_{\{x_3<0.5\}} + \mathbb{I}_{\{x_3 \geq 0.5\}}(1+2\sin(30(x_3-0.5)))], \\
		f_2(\bm{x}) &= 2 \sin(\pi x_1 x_2) + 2 x_4 - 0.8 x_5, \\
		& \qquad \qquad  + 8(x_3-0.5)^2[\mathbb{I}_{\{x_3 \geq 0.5\}} + 1.5\mathbb{I}_{\{x_3<0.5\}}(1+\sin(20(x_3-0.5)-1.5))], \\
		f_T(\bm{x}) &= 2 \sin(\pi x_1 x_2) + 8(x_3-0.5)^2 + 2 x_4 - 1 x_5,
	\end{split}
	\label{eq:friedman}
\end{align}
where $f_T$ is the original Friedman function in \cite{Friedman1991}. As before, local transfer is captured here: the first source mimics the target within the local region $x_3 \in [0,0.5]$, whereas the second source mimics the target within $x_3 \in [0.5,1.0]$.

We then generate training design points as follows. For the 1-d Forrester experiment, we employ equally-spaced design points over $[0,1]$, with $n_1 = n_2 = 32$ design points for source functions and $n_T = 7$ design points for the target. Here, we use $n_{\rm test} = 100$ evenly-spaced test points on the target function. For the 5-d Friedman function, we employ a Latin hypercube design \cite{mckay2000comparison} over $[0,1]^5$, with $n_1 = n_2 = 60$ design points for each source and $n_T = 15$ design points on the target. Here, we use $n_{\rm test} = 500$ randomly-sampled test points on the target. We note that, for both experiments, $n_T$ is intentionally set to be smaller than the usual rule-of-thumb sample size of $10d$ (see \cite{loeppky2009choosing}), to mimic the cost-intensive nature of the target system and thus the need for transfer learning.

\begin{figure}[!t]
	\centering
	\subfloat[1-d multi-source Forrester experiment.]{
		\includegraphics[width=0.96\linewidth]{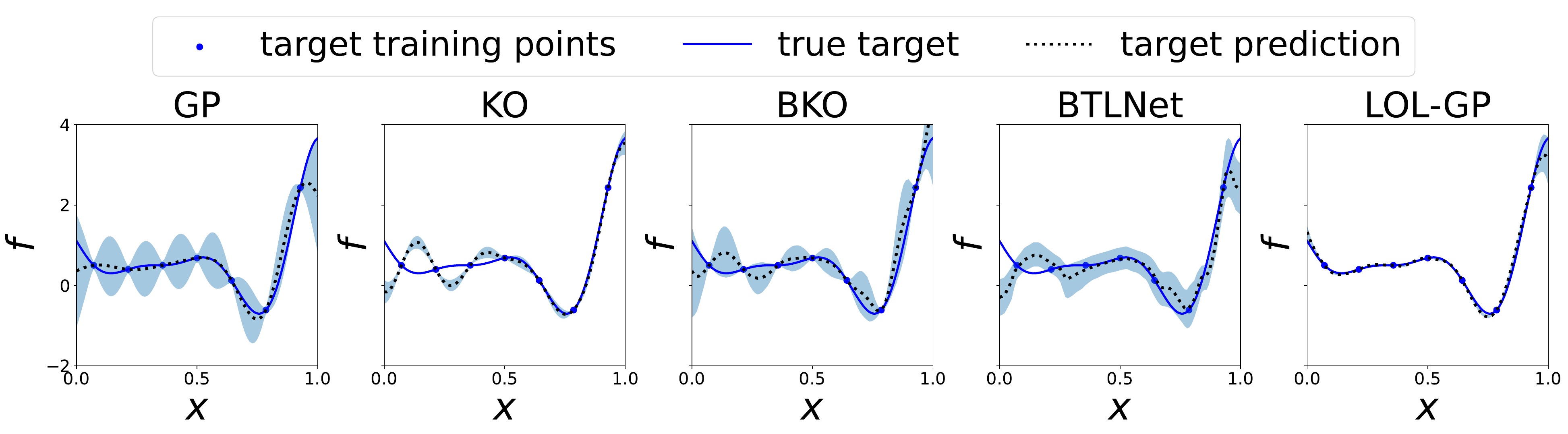}
		\label{fig: MS 1d synthetic data prediction}}
	\\
	\subfloat[5-d multi-source Friedman experiment.]{
		\includegraphics[width=0.96\linewidth]{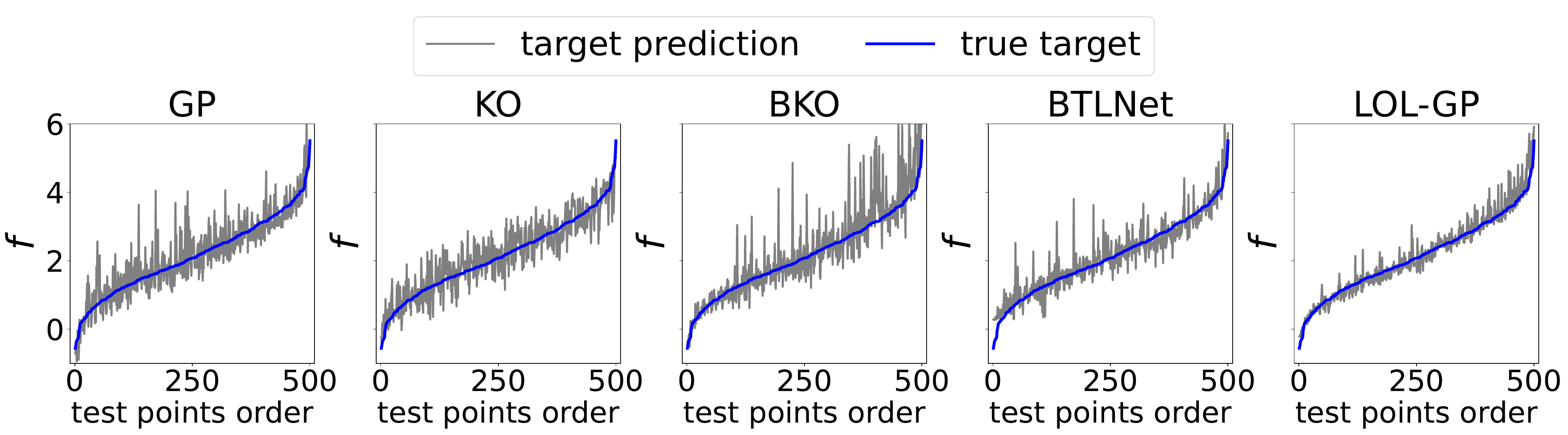}
		\label{fig: MS 5d synthetic data prediction}}
	\caption{Visualizing the predictive performance of the compared methods: (a) The true target function (solid line), its prediction (dotted line) and 95\% confidence intervals (shaded) for the 1-d multi-source Forrester experiment; (b) For the 5-d multi-source Friedman experiment, plots of the predicted test responses as a function of its true response order. The blue curve marks the true response values.}
	\label{fig: MS synthetic data prediction}
\end{figure}

\begin{table}[!t]
	\renewcommand{\arraystretch}{1.0}
	\setlength\tabcolsep{5pt}
	\setlength\abovecaptionskip{0cm}
	\caption{{Prediction metrics for the multi-source and multi-fidelity experiments. Metrics in {\color{red}{red}} indicate negative transfer (i.e., worse performance than the GP without transfer), and \textbf{bolded} metrics indicate the best performing method for each experiment.}}
	\label{tab: numerical experiments error}
	\centering
	\begin{tabular}{c cc cc cc cc}
		\toprule
		& \multicolumn{4}{c}{Multi-source} &  \multicolumn{4}{c}{Multi-fidelity} \\ \cmidrule(r){2-5}\cmidrule(r){6-9}
		& \multicolumn{2}{c}{Forrester} & \multicolumn{2}{c}{Friedman} & \multicolumn{2}{c}{Forrester} & \multicolumn{2}{c}{Branin} \\ \cmidrule(r){2-3}\cmidrule(r){4-5}\cmidrule(r){6-7}\cmidrule(r){8-9}
		& RMSE & CRPS & RMSE & CRPS & RMSE & CRPS & RMSE & CRPS \\
		\toprule
		\textbf{GP} (no transfer) &  0.307 & 0.129 & 0.474 & 0.299 & 0.249 & 0.126 & 0.138 & 0.063 \\ 
		\textbf{KO} & {\color{red}0.345} & {\color{red}0.189} & 0.441 & 0.288 & {\color{red} 0.359} & {\color{red} 0.160} & 0.137 & 0.063 \\ 
		\textbf{BKO} & 0.290 & 0.140 & {\color{red}0.613} & {\color{red}0.308} & 0.205 & 0.100 & {\color{red}0.140} & 0.058 \\
		\textbf{BTLNet} & {\color{red} 0.398} & {\color{red} 0.202} & 0.372 & 0.198 & 0.205 &  0.102 & {\color{red}0.258} & {\color{red} 0.134} \\
		\textbf{LOL-GP} & \textbf{0.096} & \textbf{0.047} & \textbf{0.274} & \textbf{0.145} & \textbf{0.180} & \textbf{0.092} & \textbf{0.084} & \textbf{0.039}\\
		\toprule
	\end{tabular}
\end{table}

\begin{figure}[!t]
	\centering
	\includegraphics[width=0.96\linewidth]{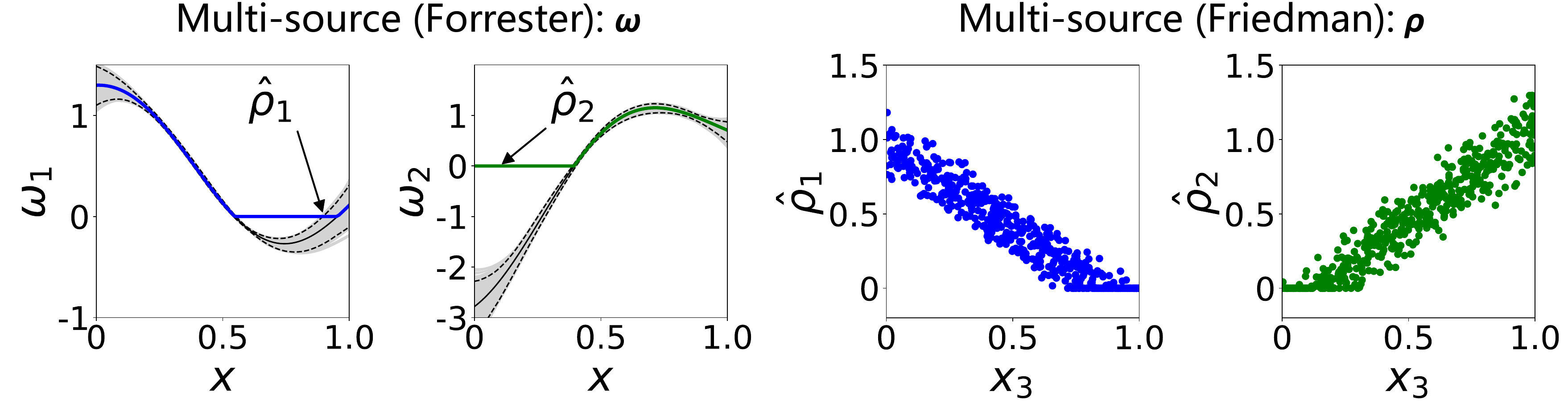}
	\caption{(Left) Visualizing the estimated latent functions $\{\omega_l({x})\}_{l=1}^2$ and its corresponding transfer functions $\{\rho_l({x})\}_{l=1}^2$ for the LOL-GP in the 1-d multi-source Forrester experiment. The posterior means and 95\% confidence intervals for $\{\omega_l({x})\}_{l=1}^2$ are marked by solid and dotted black lines, with the posterior means of $\{\rho_l({x})\}_{l=1}^2$ marked by colored lines. (Right) For the 5-d multi-source Friedman experiment, a visualization of the posterior means of the transfer functions $\{\rho_l(\bm{x})\}_{l=1}^2$ projected onto the $x_3$-space; the first is plotted in blue (left) and the second in green (right).}
	\label{fig: MS synthetic data estimated rho}
\end{figure}

Figure \ref{fig: MS 1d synthetic data prediction} visualizes the predictive performance of the compared methods for the 1-d experiment, with Table \ref{tab: numerical experiments error} reporting its predictive metrics. We see that, in the presence of local transfer, the proposed LOL-GP outperforms existing models in terms of both point prediction (i.e., lower RMSE) and probabilistic prediction (i.e., lower CRPS). From Table \ref{tab: numerical experiments error}, the KO model appears to experience considerable negative transfer: it yields worse performance than the standard GP (without transfer) for both metrics. Figure \ref{fig: MS 1d synthetic data prediction} shows that a potential reason may be an overly simplistic transfer model: by presuming constant transfer over the full domain, the KO erroneously transfers information from source 2 over [0, 0.5], which considerably worsens predictions. The LOL-GP addresses this via a careful regularization of the transfer functions $\{\rho_l({x})\}_{l=1}^2$, to identify regions where transfer is beneficial or detrimental for each source. Figure \ref{fig: MS synthetic data estimated rho} (left) visualizes this via the posterior distributions of the latent functions $\{\omega_l({x})\}_{l=1}^2$, and the posterior mean of its corresponding transfer functions $\{\rho_l({x})\}_{l=1}^2$; see Equation \eqref{eq: multi-source formulation}. We see that the latent function for source 1 is positive only within $[0,0.5]$, which is desirable as transfer is beneficial only in this local region (see Figure \ref{fig: MS synthetic data}). Conversely, the latent function for source 2 is largely positive only within $[0.5,1]$, which is desired as transfer is only beneficial in this region (see Figure \ref{fig: MS synthetic data}). The LOL-GP can thus provide robust transfer learning by identifying this underlying local transfer behavior from data. {Such analyses of $\rho_1$ and $\rho_2$ should be taken with some caution, however, due to its identifiability issues noted in Section \ref{subsec: multisystem}.}

Next, Figure \ref{fig: MS 5d synthetic data prediction} visualizes the predictive performance for the 5-d experiment, with Table \ref{tab: numerical experiments error} reporting its predictive metrics. Again, in the presence of local transfer, we see that the LOL-GP yields improved performance in terms of both point and probabilistic predictions. From Table \ref{tab: numerical experiments error}, we now see that the BKO experiences significant negative transfer: it yields worse performance for both metrics over the standard GP (without transfer). One potential reason may be the overly flexible nature of its transfer model: by permitting high flexibility for the transfer functions $\{\rho_l(\bm{x})\}_{l=1}^2$ without regularization, BKO may transfer information within regions where transfer is detrimental, which worsens performance. The LOL-GP addresses this via careful regularization of transfer functions to identify regions where transfer is beneficial or detrimental. Figure \ref{fig: MS synthetic data estimated rho} (right) visualizes this via the posterior mean of these two transfer functions, projected onto the $x_3$-space. We see that the transfer function for source 1 is largely positive only for $x_3 \in [0,0.5]$, which is desirable as transfer is beneficial only within this region (see Equation \eqref{eq:friedman}). Similarly, the transfer function for source 2 is largely positive only for $x_3 \in [0.5, 1]$, which is again desired as transfer is beneficial here. By incorporating such local transfer behavior, the LOL-GP provides robust transfer learning by tempering the risk of negative transfer. {Such analysis of $\rho_1$ and $\rho_2$ should again be taken with some caution due to potential identifiability issues.}

\subsection{Multi-fidelity transfer experiments}
\label{subsec: m-f case}

\begin{figure}[!t]
	\centering
	\subfloat[1-d multi-fidelity Forrester experiment.]{    
		\includegraphics[width=0.35\linewidth]{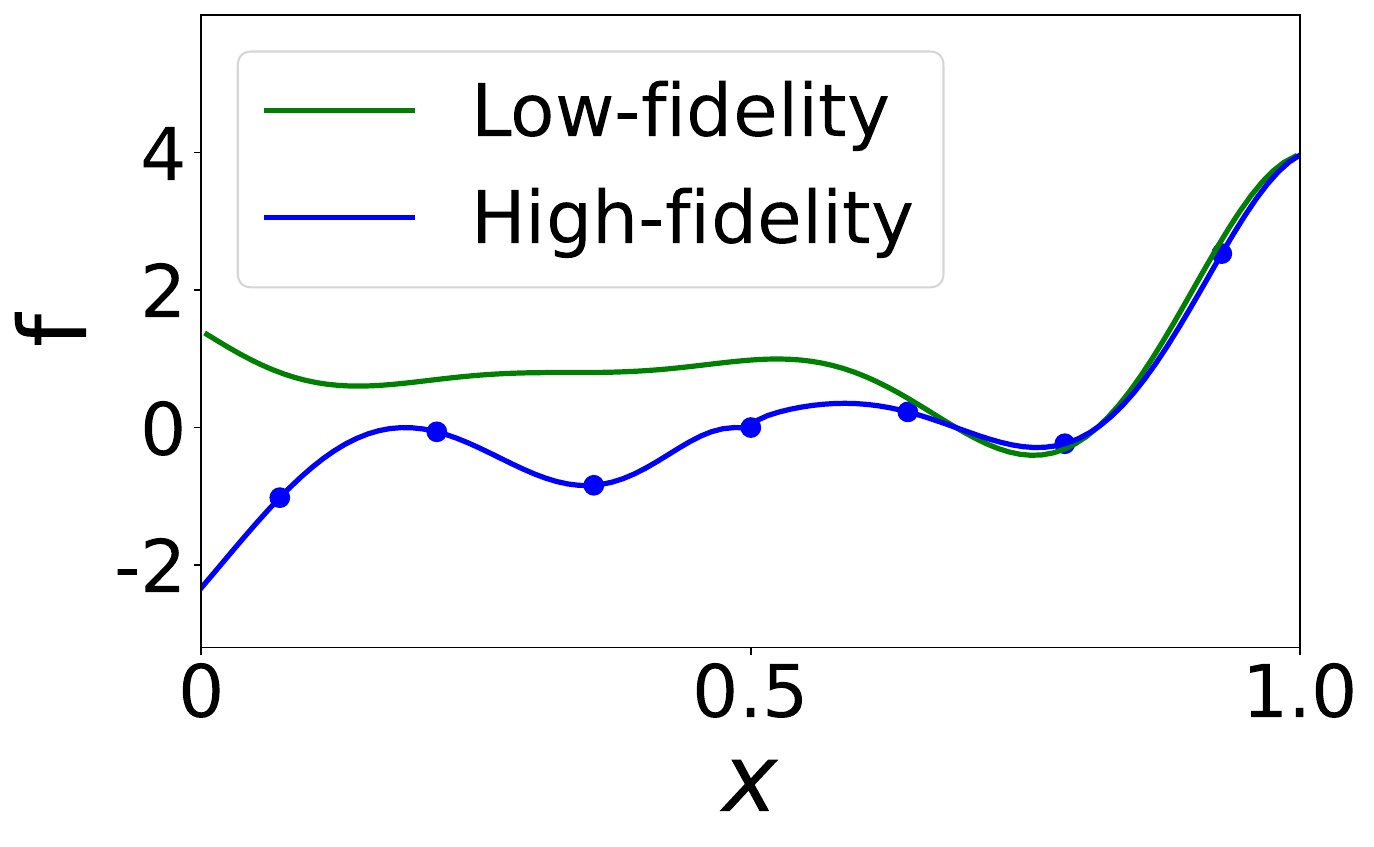}
		\label{fig: MF 1d synthetic data}}
	\subfloat[2-d multi-fidelity Branin experiment.]{    
		\includegraphics[width=0.6\linewidth]{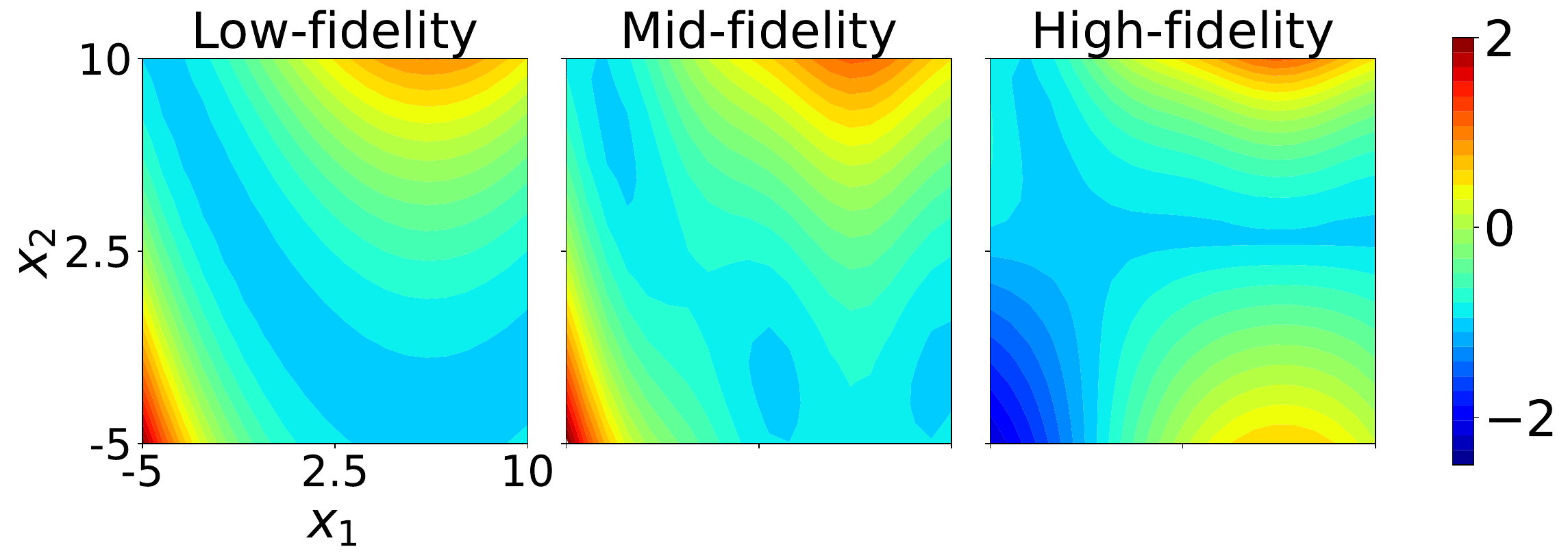}
		\label{fig: MF 2d synthetic data}}    
	\caption{Visualizing the low-fidelity (source) and high-fidelity (target) functions for the multi-fidelity experiments.}
	\label{fig: MF synthetic data}
\end{figure}

Next, we consider numerical experiments on multi-fidelity transfer. The first experiment, which again builds off of the 1-d Forrester function \cite{Forrester2008}, uses:
	\begin{align}
		\begin{split}
			&f_1(x) = 0.2(6x-2)^2 \sin(12x - 4) + 0.8, \\
			&f_2(x) = \mathbb{I}_{\{x> 0.5\}}\sqrt{2x-1}f_1(x)  + 3.2 \mathbb{I}_{\{x\leq 0.5\}}(1-2x)\left[\sin\left(14 \sqrt{x-0.5}\right)-1\right].
		\end{split}
	\end{align}
	Here, $f_1$ and $f_2$ are taken as the low-fidelity and high-fidelity functions, respectively. Figure \ref{fig: MF 1d synthetic data} visualizes these functions. As before, $f_1$ mimics $f_2$ within the local region $x \in [0.5,1]$ but not within $x \in [0,0.5]$, thus capturing local transfer from low to high fidelity. The second experiment, which builds off of the 2-d Branin function \cite{Branin1978}, uses:
	\begin{align}
		\begin{split}
			&f_1(\bm{x}) = 0.01\left(x_2 - \frac{5.1}{4\pi^2}x_1^2 + \frac{5}{\pi}x_1-1\right)^2 ,\\
			&f_2(\bm{x}) = f_1(\bm{x}) + 0.1\left[\left(1-\frac{1}{8\pi}\right)\cos(x_1^2)+1\right], \\
			&f_3(\bm{x}) = \mathbb{I}_{\{x_2>3\}}\frac{x_2-3}{7}f_2(\bm{x}) - \mathbb{I}_{\{x_2 \leq 3\}}\frac{x_2-3}{5}\sin\left(\frac{2(x_1+1)}{3\pi}\right).
		\end{split}
	\end{align}
	Here, $f_1$, $f_2$ and $f_3$ are taken as the low-, medium- and high-fidelity functions, respectively. Figure \ref{fig: MF 2d synthetic data} visualizes these functions. Note that $f_2$ mimics the high-fidelity function $f_3$ when $x_2 > 3$ but is considerably different otherwise, which captures local transfer.

{We note that the above multi-fidelity experiments are designed to capture the intuition that high-fidelity simulators typically capture finer-scale (i.e., more complex) structure than lower-fidelity simulators. This can be seen from Figure \ref{fig: MF synthetic data}: the high-fidelity Forrester function is noticeably more ``wiggly'' within $[0,0.5]$, and the high-fidelity Branin function has greater complexity when $x_2 \leq 3$. This intuition of more complex structure on the target, however, might not extend to the earlier multi-source setting. There, while one has more data on source systems than the target, the target system may not be more expensive nor more complex than source systems. Take, e.g., the transfer learning application in \cite{TransferSurrogate-HeavyIon2022}, where the goal is to emulate particle collision observables under a target particlization system. While there exists existing data from prior studies on related (i.e., source) particlization systems, such systems are as sophisticated as the target system and require similar simulation times.}

We then generate design points as follows. For the 1-d experiment, we take equally-spaced design points on $[0,1]$, with $n_1 = 21$ and $n_2 = 7$; the high-fidelity (target) design are nested within the low-fidelity design. Here, $n_{\rm test} = 100$ equally-spaced points are used for testing. For the 2-d experiment, design points are selected via a sliced Latin hypercube design \cite{SLHD2015} (first generated on $[0,1]^2$, then rescaled to $[-5,10]^2$), with sample sizes $n_1 = 48$, $n_2 = 24$ and $n_3 = 12$; such points are selected in a nested fashion. Here, $n_{\rm test} = 400$ grid points are used for testing.

\begin{figure}[!t]
	\centering
	\subfloat[1-d multi-fidelity Forrester experiment.]{
		\includegraphics[width=0.96\linewidth]{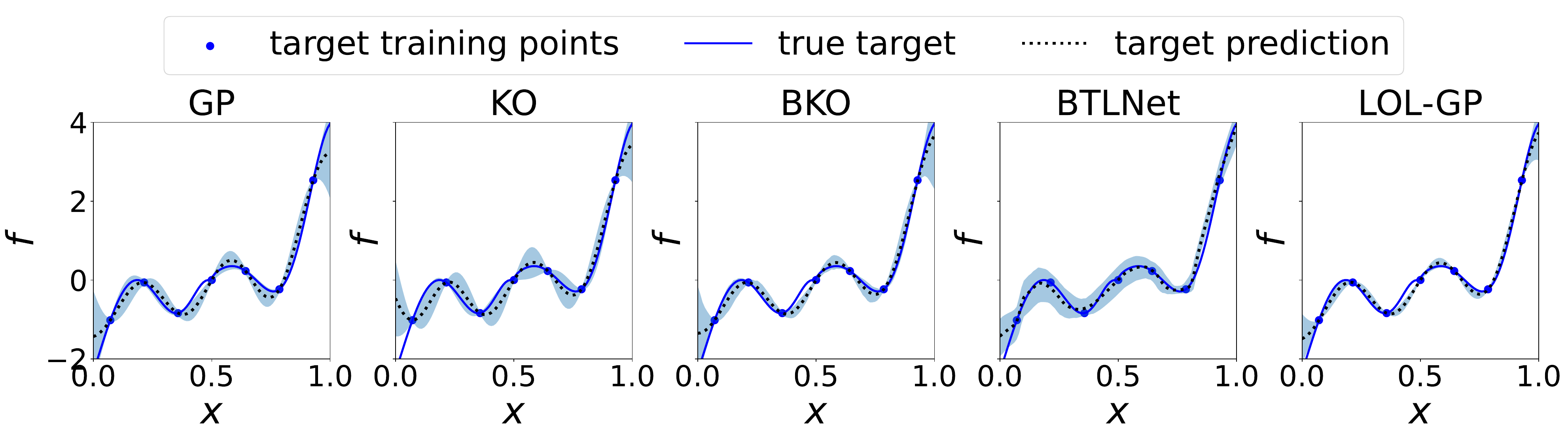}
		\label{fig: MF 1d synthetic data prediction}}
	\\
	\subfloat[2-d multi-fidelity Branin experiment.]{
		\includegraphics[width=0.96\linewidth]{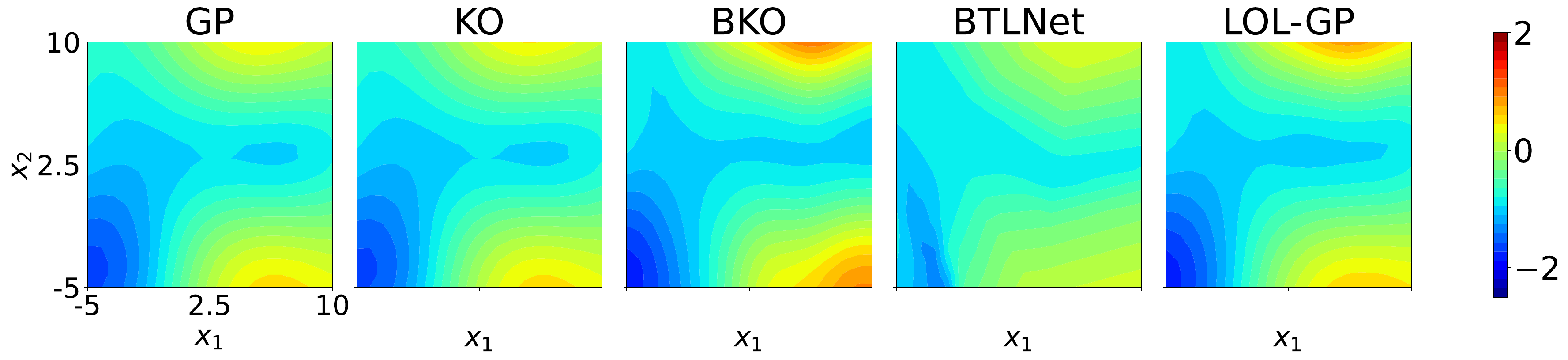}
		\label{fig: MF 2d synthetic data prediction}}
	\caption{Visualizing the predictive performance of the compared methods. (a) The true high-fidelity function (solid line), its prediction (dotted line) and 95\% confidence intervals (shaded) for the 1-d multi-fidelity Forrester experiment. (b) Contours of the predicted high-fidelity functions for the 2-d multi-fidelity Branin experiment.}
	\label{fig: MF synthetic data prediction}
\end{figure}

\begin{figure}[!t]
	\centering
	\includegraphics[width=0.65\linewidth]{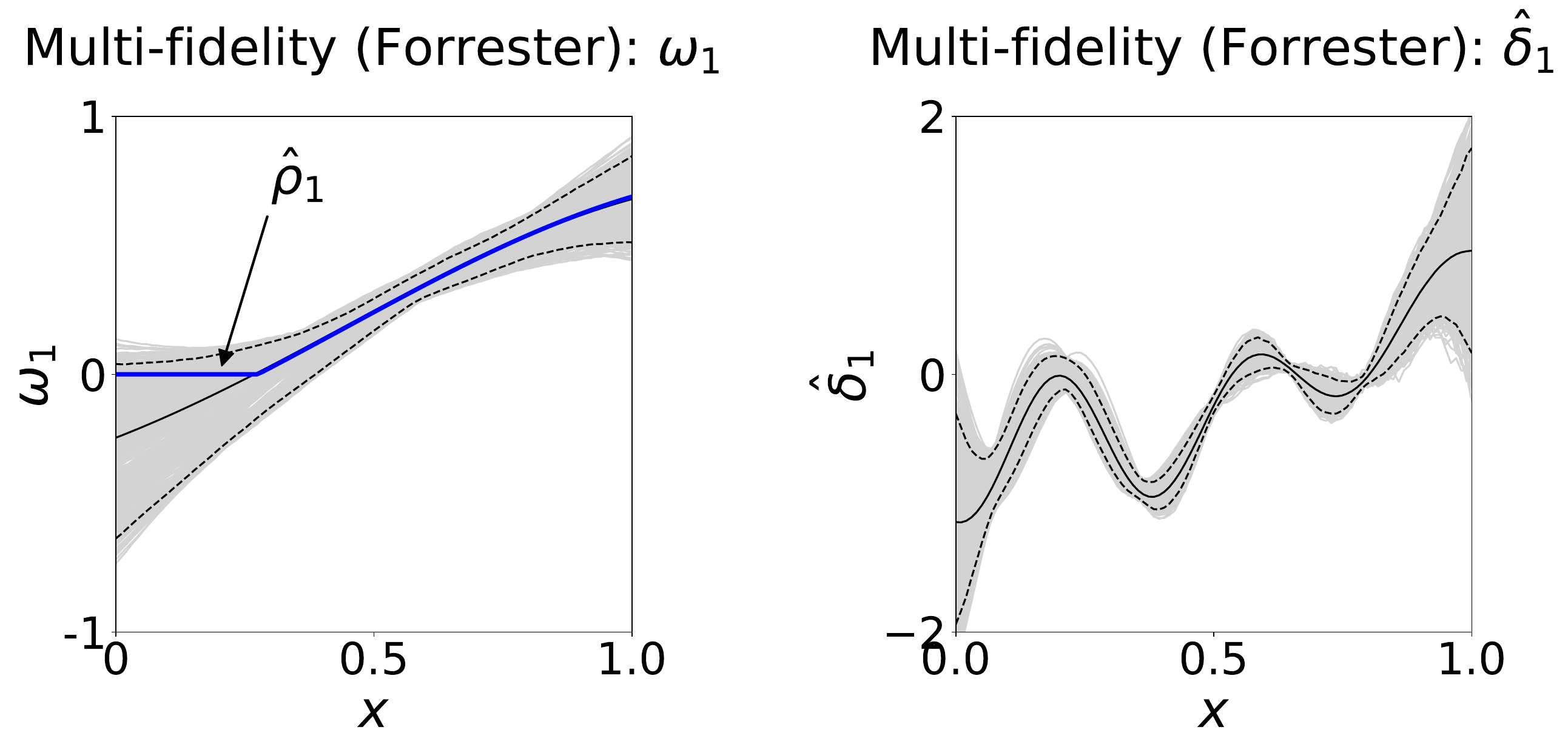}
	\caption{(Left) Visualizing the estimated latent function $\omega_1(x)$ and its corresponding transfer function $\rho_1(x)$ for the LOL-GP in the 1-d multi-fidelity Forrester experiment. The posterior means and 95\% confidence intervals for $\omega_1(x)$ are marked by solid and dotted black lines, and the posterior means for $\rho_1(x)$ is marked by the blue line. (Right) Visualizing the estimated discrepancy function $\delta_1(x)$ for the LOL-GP in the 1-d multi-fidelity Forrester experiment. Its posterior means and 95\% confidence intervals are marked by solid and dotted black lines.}
	\label{fig: MF synthetic data estimated rho}
\end{figure}

\begin{figure}[!t]
	\centering
	\includegraphics[width=0.7\linewidth]{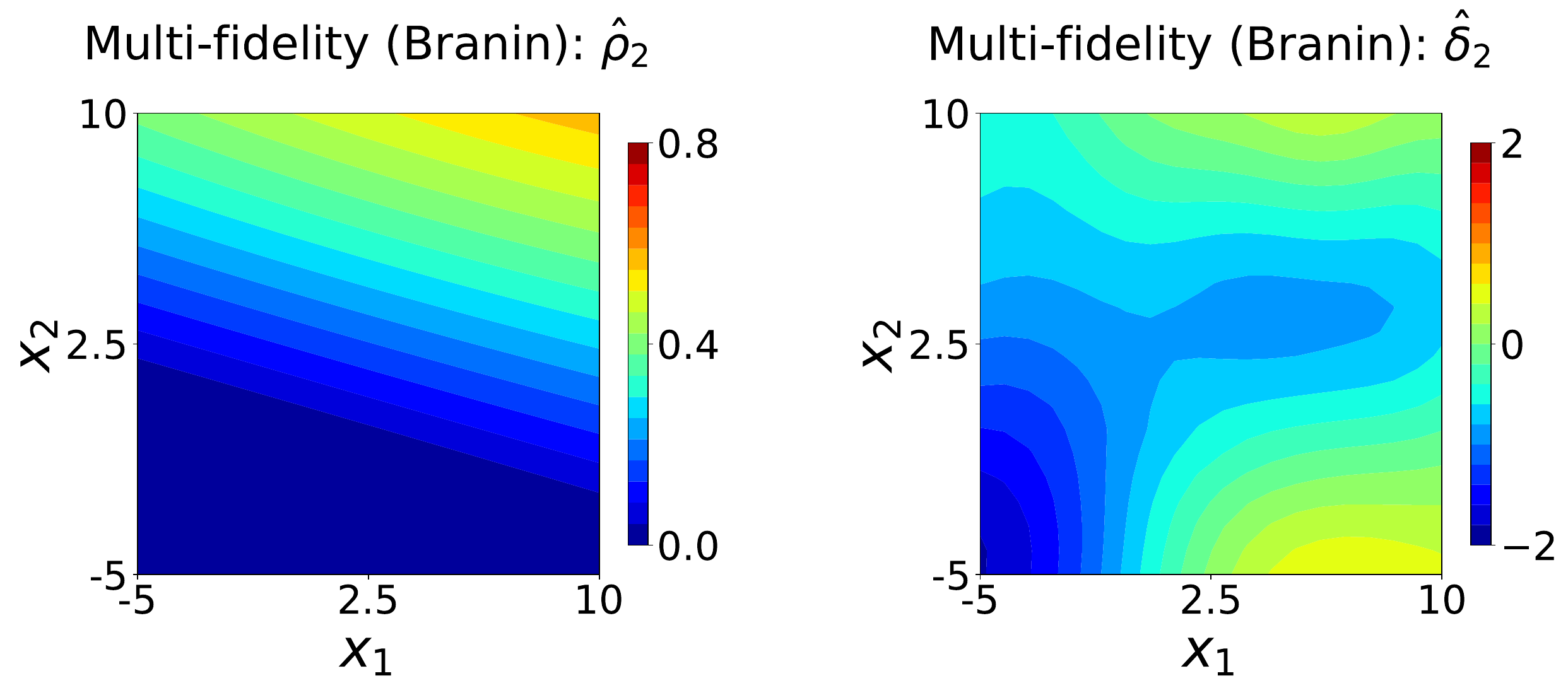}
	\caption{(Left) A contour plot of the posterior mean for the transfer function $\rho_2(\bm{x})$ in the 2-d multi-fidelity Branin experiment. (Right) A contour plot of the posterior mean for the discrepancy function $\delta_2(\bm{x})$ in the 2-d multi-fidelity Branin experiment.}
	\label{fig: MF synthetic data estimated delta}
\end{figure}

Figure \ref{fig: MF 1d synthetic data prediction} visualizes the predictive performance for the 1-d experiment, with Table \ref{tab: numerical experiments error} reporting its predictive metrics. Again, in the presence of local transfer, the LOL-GP yields better performance over existing methods for both point predictions (i.e., lower RMSE) and probabilistic predictions (i.e., lower CRPS). {Figure \ref{fig: MF synthetic data estimated rho} shows the posterior distributions of the latent and transfer functions $\omega_1(x)$ and $\rho_1(x)$ for the LOL-GP, as well as that for its discrepancy function $\delta_1(x)$. Within the region $[0.5,1]$, we see the posterior mean of the transfer function $\rho_1(x)$ is largely positive and the posterior mean of the discrepancy is close to zero, which is not surprising as transfer is expected to be beneficial only within this region (see Figure \ref{fig: MF 1d synthetic data}). Again, such analyses should be taken with some caution due to its potential identifiability issues.}

Next, Figure \ref{fig: MF 2d synthetic data prediction} shows the predictive performance for the 2-d experiment, with Table \ref{tab: numerical experiments error} summarizing its metrics. The LOL-GP again outperforms existing methods in the presence of local transfer. {Here, KO performs comparably to the standard GP without transfer, and BKO and BTLNet experience negative transfer in that their metrics are worse than the standard GP. This hints at the dangers of an overly complex transfer model; with limited target data, such models can deteriorate performance and induce negative transfer.} {Figure \ref{fig: MF synthetic data estimated delta} shows the posterior distributions for the transfer function ${\rho}_2(\bm{x})$ and discrepancy function ${\delta}_2(\bm{x})$. We see that, with the region $x_2 \in [2, 10]$,  the posterior mean of ${\rho}_2(\bm{x})$ is largely positive and the posterior mean of ${\delta}_2(\bm{x})$ is close to zero; this is not surprising as transfer is expected to be beneficial within this region (see Figure \ref{fig: MF 2d synthetic data}).} By incorporating such local transfer, the LOL-GP facilitates robust transfer learning with limited target data.

It is also worth investigating, for the compared models, the computation times for model fitting (both for hyperparameter optimization and Gibbs sampling, if needed) and the efficiency of their MCMC samplers (if needed). For the latter, we report its effective sample size (ESS; \cite{vats2019multivariate}), which measures the quality of an MCMC chain relative to i.i.d. Monte Carlo sampling; a larger ESS indicates greater effectiveness of the MCMC sampler at reducing autocorrelation. Table \ref{tab: mf fitting time} summarizes the computation times and ESS for the above multi-fidelity experiments. On computation time, the standard GP (without transfer) and the KO models have the quickest times, which is not surprising as such approaches do not require MCMC. The LOL-GP requires noticeably longer computation, which is expected as it needs to sample a greater number of latent parameters to model local transfer. On MCMC efficiency, we see that BKO has slightly higher ESS than LOL-GP, which suggests its MCMC sampler has higher efficiency. One plausible reason for the lower ESS of the LOL-GP is that its identifiability issues (see Section \ref{subsec: multisystem}) may induce higher correlations in the posterior distribution, which reduce MCMC sampling efficiency. We thus recommend ESS checks along with standard MCMC diagnostics \cite{gelman1995bayesian} to ensure adequate MCMC mixing for our model.

\begin{table}[!t]
	\centering
	\begin{tabular}{ccccccc}
		\toprule
		& & \makecell{\textbf{GP} \\ (no transfer)} & \textbf{KO} & \textbf{BKO} & \textbf{BTLNet} & \textbf{LOL-GP}\\ 
		\toprule
		\multirow{2}{*}{1-d multi-fidelity Forrester} & {Time} & 0.2 & 0.6 & 4 & 140 & 203 \\ 
		& {ESS} & - & - & 936 & - & 852 \\
		\hline
		\multirow{2}{*}{2-d multi-fidelity Branin} & {Time} & 0.2 & 0.9 & 8 & 145 & 1670 \\
		& {ESS} & - & - & 828 & - & 640 \\
		\toprule
	\end{tabular}
	\caption{{Computation times for model fitting (in seconds, both for hyperparameter optimization and Gibbs sampling if needed) and effective sample sizes (ESS; \cite{vats2019multivariate}) for the 1-d multi-fidelity Forrester and 2-d multi-fidelity Branin experiments. Here, ESS is reported only for methods requiring MCMC, and is computed using an MCMC chain with 9,000 iterations (after burn-in).}}
	\label{tab: mf fitting time}
\end{table}

{The above experiments presume the true source and target systems behave similarly in certain regions of the parameter space but differently in others. While this property arises broadly in the physical sciences (see the Introduction) and is the primary focus of this work, it is useful to investigate how the LOL-GP performs when transfer is desirable over the full domain. The next experiment explores this setting. We use the following modification of the 1-d Forrester function:
	\begin{equation}
		\label{eq:mfnotransfer}
		\begin{split}
			&f_1(x) = 0.2(6x-2)^2 \sin(12x - 4) + 0.5,\\
			&f_2(x) = 0.2(6x-2)^2 \sin(12x - 4) + 1.6(1 - x)+ 0.5,
		\end{split}
	\end{equation}
	where $f_1$ and $f_2$ are taken as the low-fidelity and high-fidelity functions, respectively. Note that these two functions have similar trends over the full parameter space $[0,1]$, thus transfer may be beneficial throughout. Table \ref{tab: multi-fidelity emulation error without local transfer} summarizes the predictive metrics for each model. Here, all models except BTLNet significantly outperform the standard GP model without transfer, with the KO, BKO and LOL-GP having similar metrics. This is not too surprising: when transfer is desirable over the full domain, the LOL-GP can account for this via a positive fit of the latent function $\omega_1(x) > 0$, resulting in comparable performance with existing models.
	
	\begin{table}[!t] 
		\renewcommand{\arraystretch}{1.1}
		\setlength\tabcolsep{5pt}
		\setlength\abovecaptionskip{0cm} 
		\caption{Prediction metrics for the multi-fidelity Forrester experiment \eqref{eq:mfnotransfer} without local transfer. Metrics in {\color{red}{red}} indicate negative transfer, and \textbf{bolded} metrics indicate the best performing method. }
		\label{tab: multi-fidelity emulation error without local transfer}
		\centering
		\begin{tabular}{c ccccc}
			\toprule
			& \makecell*[c]{\textbf{GP} (no transfer)} &  \makecell*[c]{\textbf{KO}}  & \makecell*[c]{\textbf{BKO}} & \makecell*[c]{\textbf{BTLNet}} & \makecell*[c]{\textbf{LOL-GP}} \\
			\toprule
			RMSE & 0.0879 & \textbf{0.0121} & 0.0122 & {\color{red} 0.131} & 0.0122 \\
			CRPS & 0.0296 & {0.0100} & \textbf{0.0100} & {\color{red} 0.078} & {0.0100} \\
			\toprule
		\end{tabular}
	\end{table}
}

\section{Surrogate Modeling of Stress Analysis in Jet Engine Turbines}
\label{sec: stress emulation}

Jet engines are broadly used in commercial aviation applications \cite{narayanan2023physics}, and its reliable performance is of upmost importance during operation. Such engines generate thrust as follows. A heated gas is sucked into and passed to a compressor, where it is subjected to high pressure and high temperatures. This gas is then discharged as a fast-moving jet through a turbine, which then generates the necessary thrust for propulsion. Here, a careful stress analysis of turbine blades is essential for assessing engine reliability; these blades should have enough strength to withstand any stress and/or deformations in such a high-pressure (up to 0.7 MPa) and high-temperature (up to 1050 ${}^{\circ}$C) environment.

\begin{figure}[!b]
	\centering
	\includegraphics[width=0.60\linewidth]{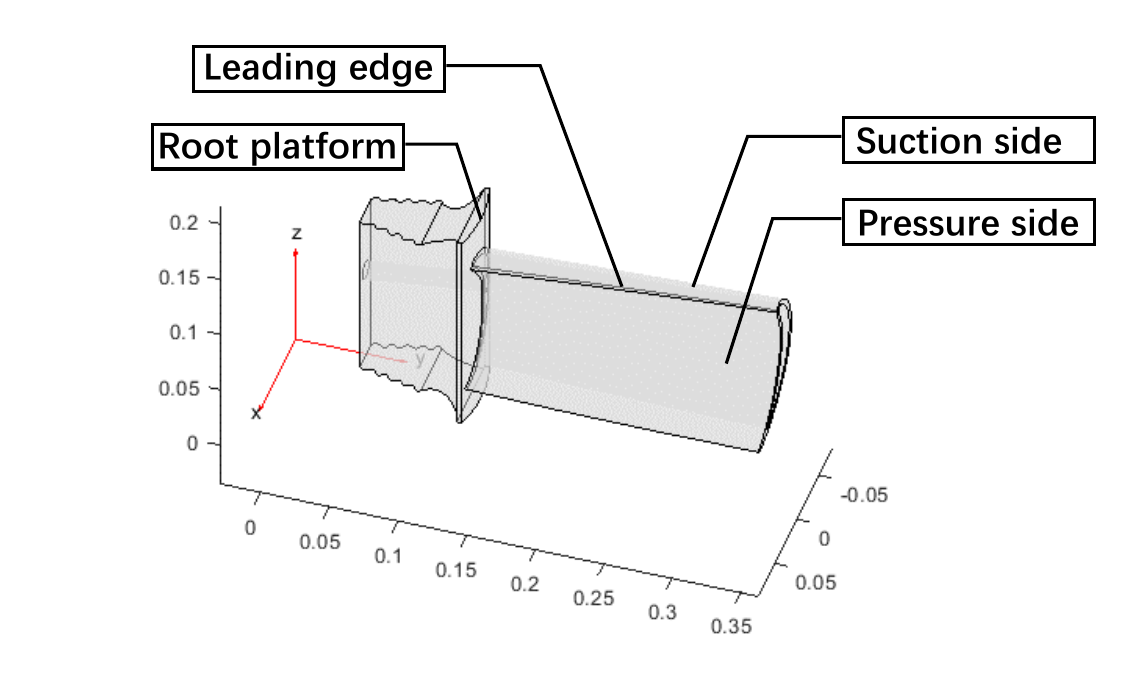}
	\caption{Schematic of the considered jet engine turbine blade in \cite{matlab_jet}.}
	\label{fig: blade structure}
\end{figure}

Figure \ref{fig: blade structure} shows the blade schematic for the considered jet engine turbine in our study, following \cite{matlab_jet}. For initial analysis, physical experiments can be prohibitively expensive, as it requires considerable costs for engine prototyping, manufacturing and subsequent testing. We thus adopt a virtual experimentation approach, which uses finite element analysis (FEA) for simulating the stress profile of turbine blades during operation. For this, we make use of the MATLAB module in \cite{matlab_jet}, which can perform such a simulation under different operating environments and blade materials. This module leverages a static-structural model on turbine blades, coupled with steady-state thermal model for thermal expansion; further details are provided in \cite{matlab_jet}. The underlying system of governing equations is then solved via a mesh discretization of the spatiotemporal domain.

\begin{figure}[!t]
	\centering
	\includegraphics[width=0.75\linewidth]{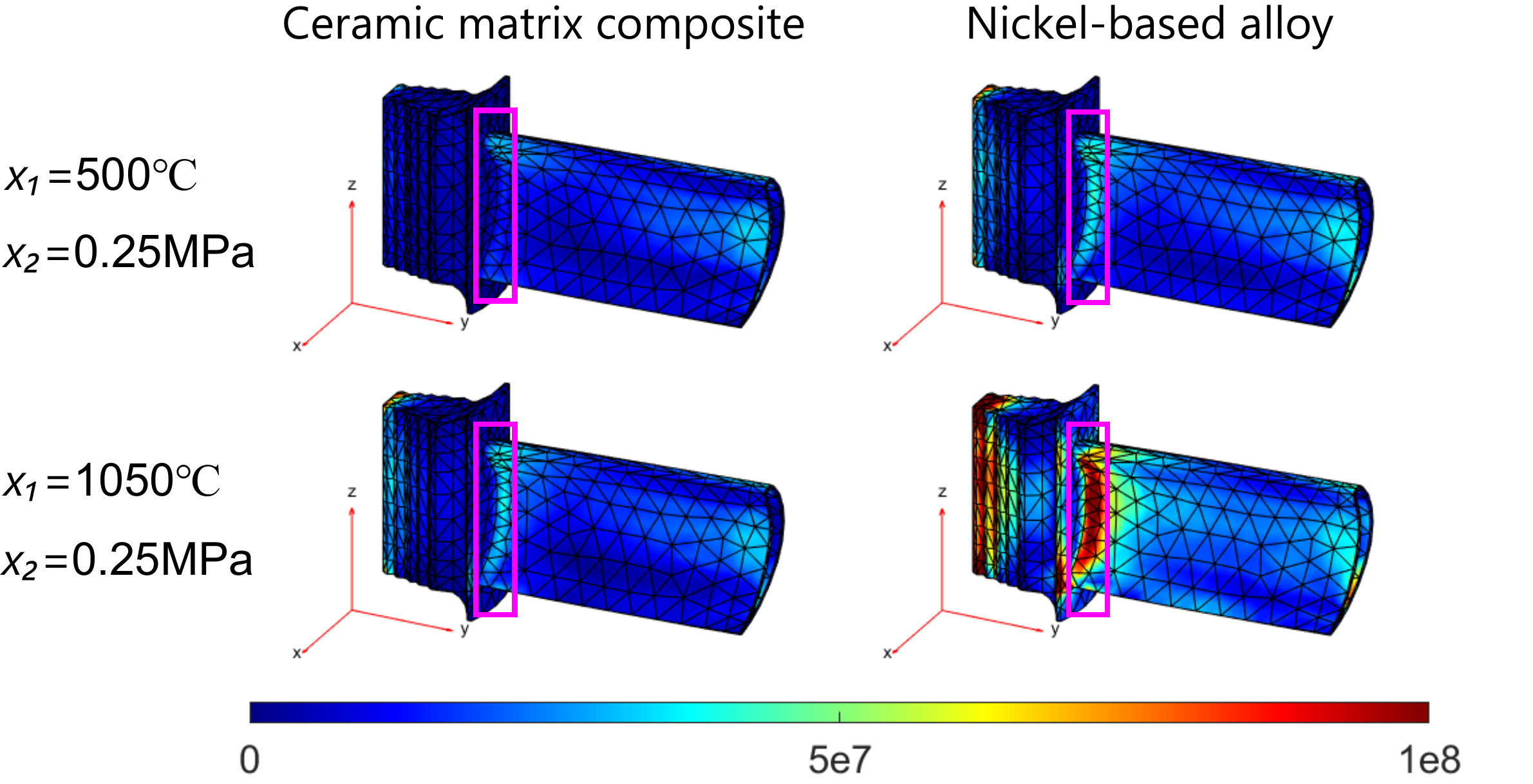}
	\caption{Visualizing the simulated stress solution in the multi-source set-up for our jet turbine application. The left and right columns correspond to ceramic matrix composite (source) and nickel-based alloy (target) blades, and the top and bottom rows correspond to different parameter settings for $(x_1, x_2)$. The response of interest is the maximum stress at the root of the blade, which is marked by the pink rectangle.}
	\label{fig: Blade MS stress solution}
\end{figure}

With this, we explore two set-ups for surrogate modeling, the first for multi-source transfer and the second for multi-fidelity transfer. For both set-ups, the response of interest is the maximum simulated stress at the root of the blade. For the multi-source set-up, we use a single source system (i.e., $L=1$). The source and target systems employ different blade material: the source uses ceramic matrix composites \cite{Ceramic2014}, whereas the target uses nickel-based alloys \cite{SuperAlloys2002}. Here, $d=2$ input parameters (controlling operating conditions) are considered: the pressure load on the pressure side (ranging from $[0.25,0.75]$ MPa), and the gas temperature (ranging from $[550, 1050]$ $^{\circ}$C). Figure \ref{fig: Blade MS stress solution} shows the simulated stress profiles using these different blade materials at two choices of input parameters. We see some evidence for local transfer: for the first parametrization, the source and target systems are quite similar, whereas for the second parametrization, the two systems are markedly different.

\begin{figure}[!t]
	\centering
	\includegraphics[width=0.98\linewidth]{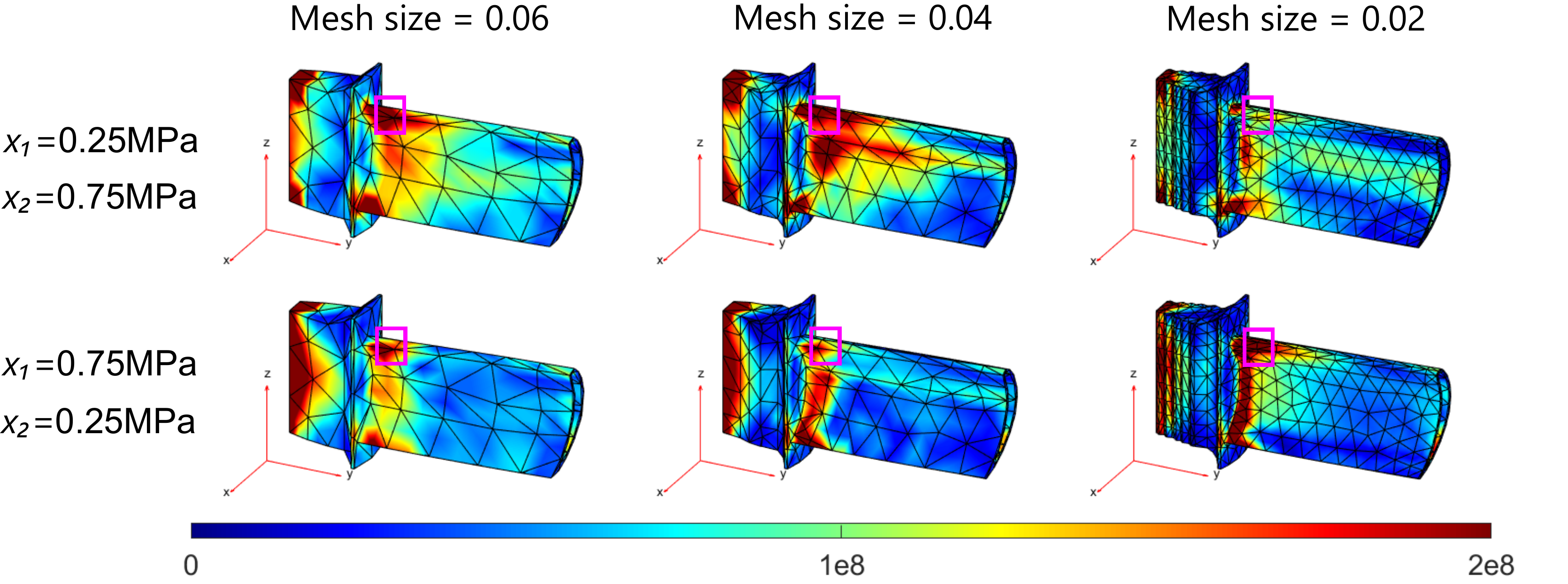}
	\caption{Visualizing the simulated stress solution in the multi-fidelity set-up  for our jet turbine application. Different columns show the simulations at different mesh sizes, and the top and bottom rows correspond to different parameter settings for $(x_1,x_2)$. The response of interest is the maximum stress at the root of the blade, which is marked by the pink rectangle.}
	\label{fig: Blade MF stress solution}
\end{figure}

For the multi-fidelity set-up, we employ $L=2$ lower-fidelity systems with maximum mesh size of 0.06 and 0.04; the high-fidelity target system uses a smaller maximum mesh size of 0.02. Here, $d=2$ input parameters (controlling operating conditions) are considered: the pressure loads on the pressure side and on the suction side (both ranging from $[0.25,0.75]$ MPa). The pressure and suction sides of the blade are labeled in Figure \ref{fig: blade structure}. Figure \ref{fig: Blade MF stress solution} shows the simulated stress profiles with varying fidelities at two choices of input parameters. Again, we see evidence for local transfer: at the first input, the systems are considerably different from low to high fidelity, whereas at the second input, they are quite similar. For brevity, we refer to the two considered input parameters as $x_1$ and $x_2$ for both the multi-source and the multi-fidelity set-up. 

\begin{figure}[!t]
	\centering
	\subfloat[Multi-source set-up.]{
		\includegraphics[width=0.395\linewidth]{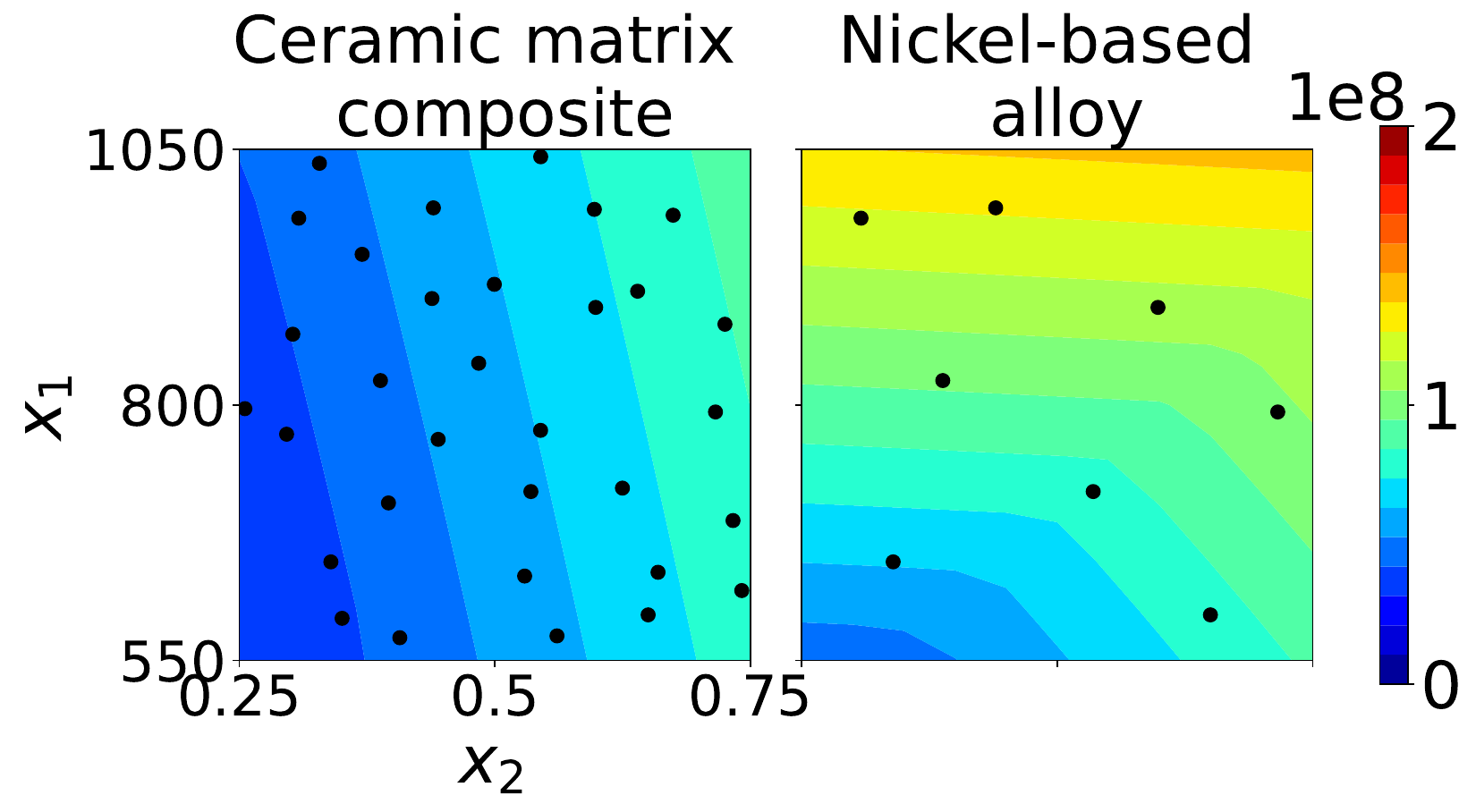}
		\label{fig: Blade MS data}}
	\subfloat[Multi-fidelity set-up.]{
		\includegraphics[width=0.57\linewidth]{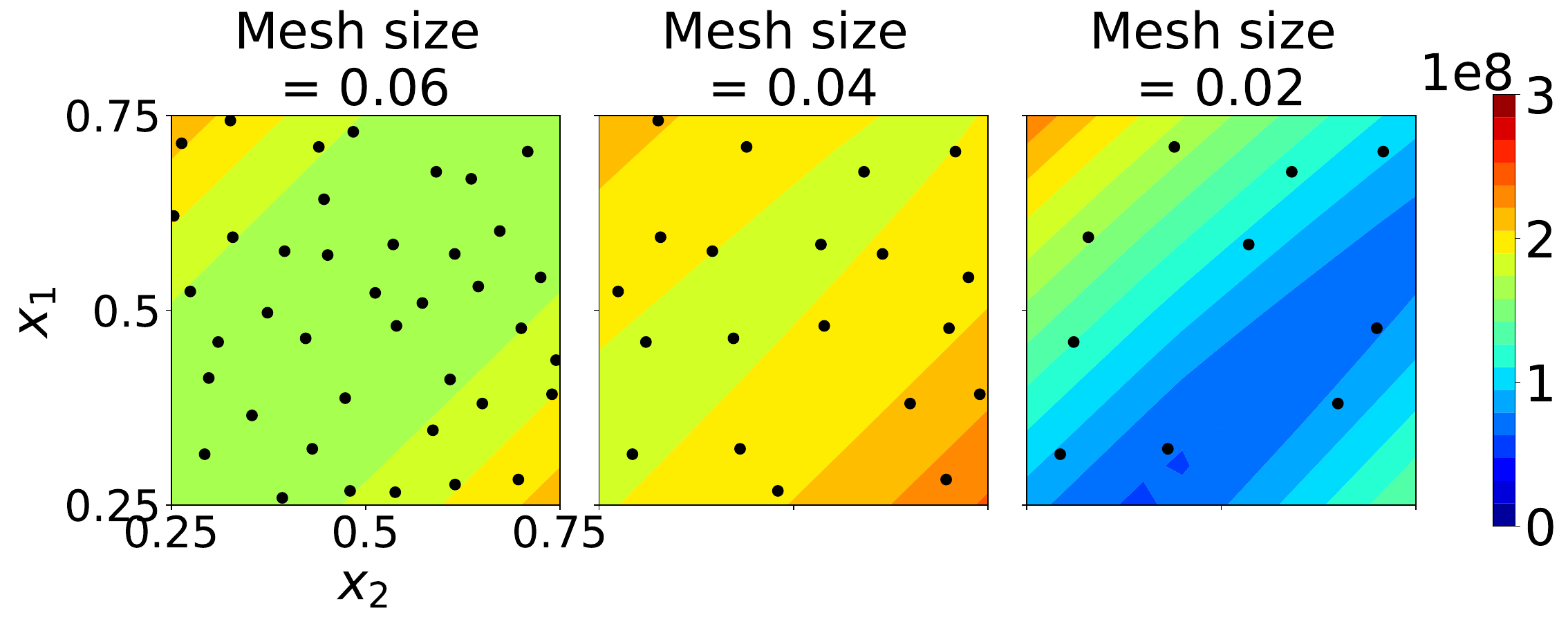}
		\label{fig: Blade MF data}}
	\caption{Visualization of the true source and target response surfaces in the multi-source and multi-fidelity set-ups for our jet turbine application. Design points are marked by black dots.}
	\label{fig: Blade data}
\end{figure}

We then generate design points as follows. For the multi-source set-up, we take $n_1 = 32$ design points on the source and $n_T = 8$ points on the target, both generated via separate Latin hypercube designs. For the multi-fidelity set-up, we take $n_1 = 40$, $n_2 = 20$ and $n_3 = 10$ design points on the low-, medium- and high-fidelity simulators. Such points are obtained via a sliced Latin hypercube design \cite{SLHD2015}, with design points nested for increasing fidelity levels. For both set-ups, $n_{\rm test} = 121$ separate grid points are used for testing. Figure \ref{fig: Blade data} shows the training design points for both set-ups, along with its corresponding response surfaces.

\begin{figure}[!t]
	\centering
	\includegraphics[width=0.98\linewidth]{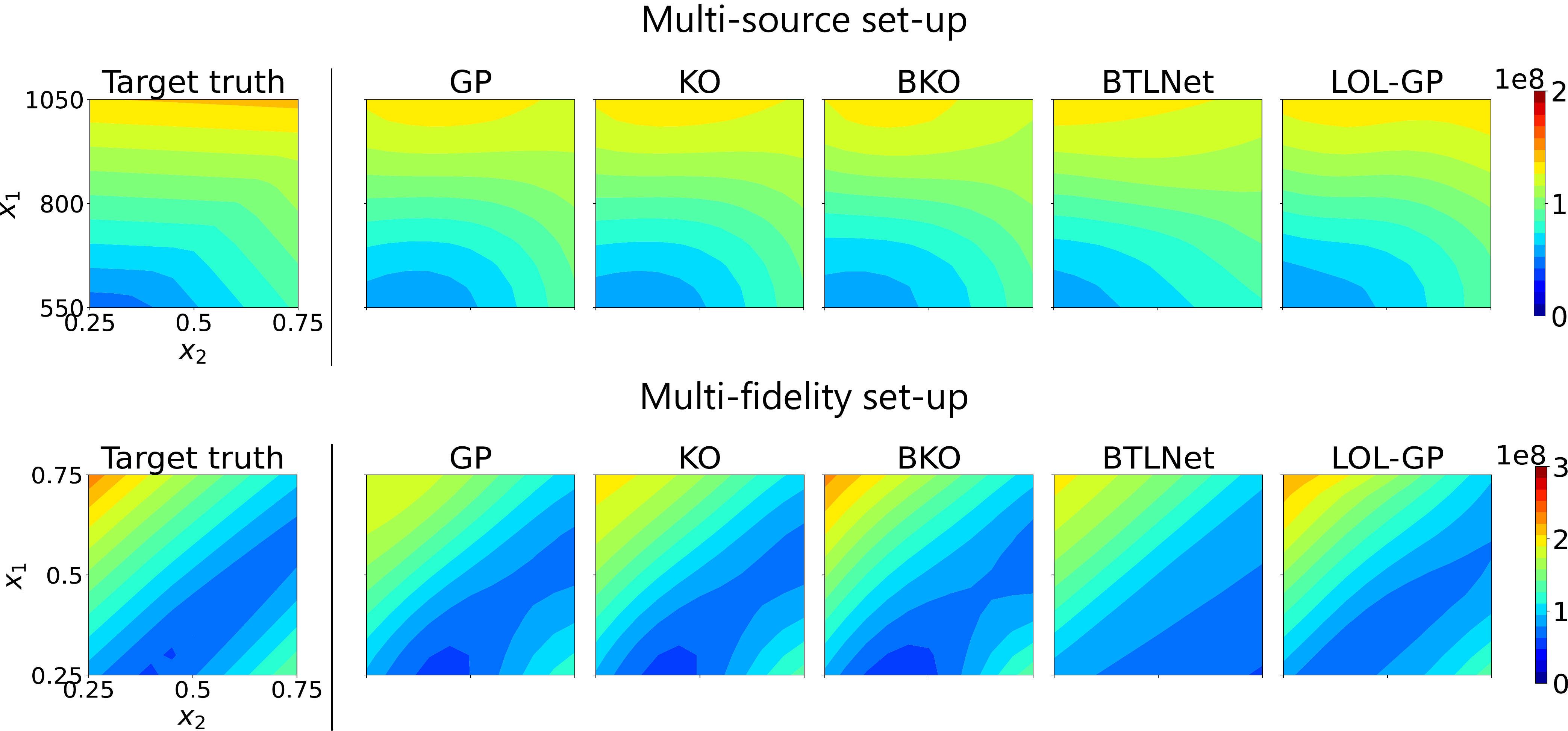}
	\caption{Visualizing the true target response surface (left), and the predicted response surfaces (right) from the compared surrogates for our jet turbine application.}
	\label{fig: Blade prediction}
\end{figure}

\begin{figure}[!t]
	\centering
	\includegraphics[width=0.80\linewidth]{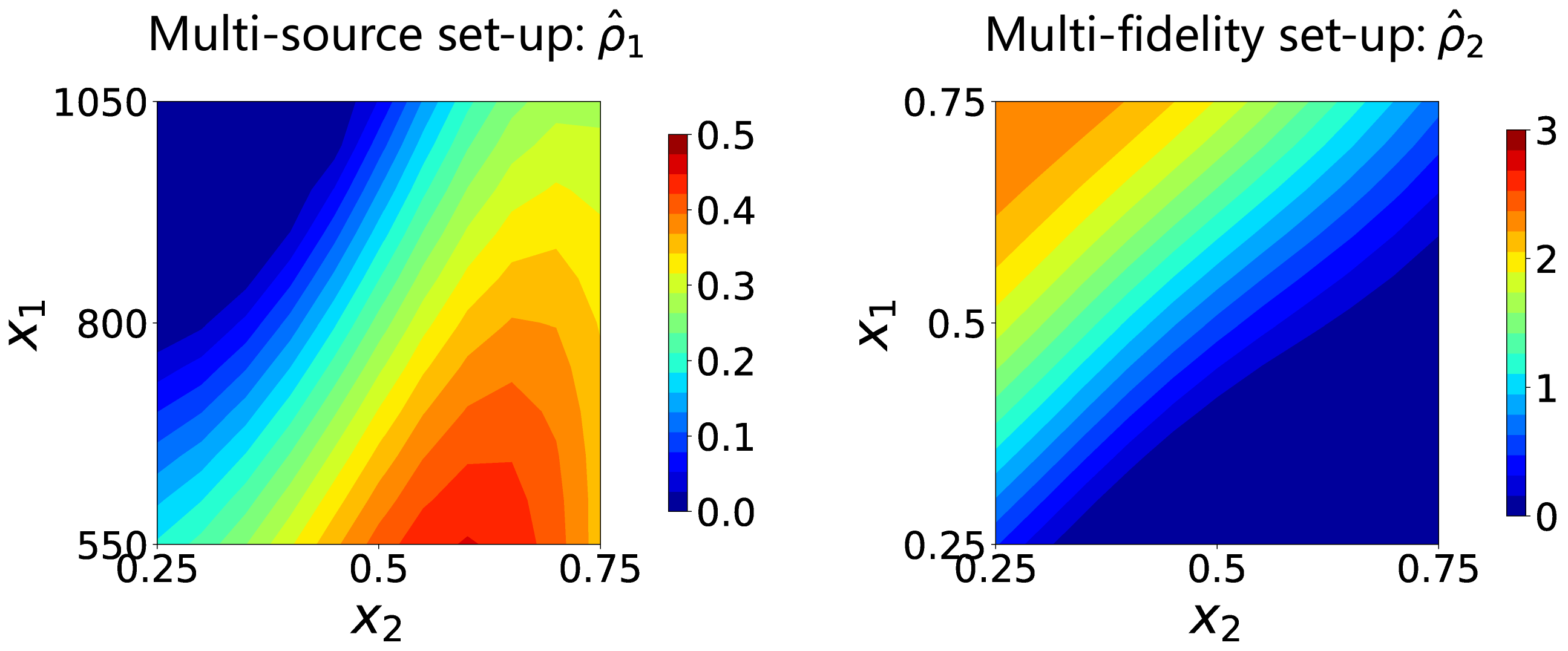}
	\caption{(Left) Visualizing the posterior mean of the transfer function $\rho_1(\bm{x})$ for the LOL-GP in the multi-source set-up  for our jet turbine application. (Right) Visualizing the posterior mean of the transfer function $\rho_2(\bm{x})$ for the LOL-GP in the multi-fidelity set-up. }
	\label{fig: Blade rho estimate}
\end{figure}

Consider first the multi-source set-up, with predictive metrics in Table \ref{tab: stress emulation error} and visualization in Figure \ref{fig: Blade prediction}. As before, the proposed LOL-GP yields the best predictive performance in terms of both RMSE and CRPS. Here, all three existing models exhibit negative transfer: its RMSE and/or CRPS can be considerably larger than that for the GP surrogate with no transfer. The LOL-GP appears to mitigate this by leveraging our prior observation of local transfer, i.e., that ceramic blades (source) may perform similarly to nickel-based blades (target) at certain parametrizations but not at others. To explore this, Figure \ref{fig: Blade rho estimate} (left) shows the estimated transfer function $\rho(\bm{x})$ for the LOL-GP. We see that this transfer function is near zero in the top-left corner where $x_1$ (gas temperature) is high and $x_2$ (pressure load on the pressure side) is low. This corroborates our earlier observation of local transfer from Figure \ref{fig: Blade MS stress solution}: for a parameter setting with $x_1$ high and $x_2$ low, the simulated stress can vary considerably with different blade materials. Our model identifies and incorporates this local transfer behavior to facilitate \textit{robust} transfer learning with limited target data. {As before, such analyses should be taken with some caution due to the identifiability issues noted in Section \ref{subsec: multisystem}.}

\begin{table}[!t] 
	\renewcommand{\arraystretch}{1.0}
	\setlength\tabcolsep{5pt}
	\setlength\abovecaptionskip{0cm}  
	\caption{Prediction metrics for surrogate modeling of the jet turbine application. Metrics in {\color{red}{red}} indicate negative transfer (i.e., worse performance than the GP without transfer), and \textbf{bolded} metrics indicate the best performing method for each experiment.}
	\label{tab: stress emulation error}
	\centering
	\begin{tabular}{cc ccccc}
		\toprule
		& & \makecell*[c]{\textbf{GP} (no transfer)} &  \makecell*[c]{\textbf{KO}}  & \makecell*[c]{\textbf{BKO}} & \makecell*[c]{\textbf{BTLNet}} & \makecell*[c]{\textbf{LOL-GP}} \\
		\toprule
		\multirow{2}{*}{Multi-source} & RMSE & 0.131 & 0.118 & {\color{red} 0.177} & {\color{red}0.166} & \textbf{0.100} \\
		& CRPS & 0.062 & {\color{red}{0.065}} & {\color{red}0.087} & {\color{red}0.092} & \textbf{0.053} \\     
		\hline
		\multirow{2}{*}{Multi-fidelity} & RMSE & 0.166 & 0.127 & 0.106 & {\color{red}0.342} & \textbf{0.078} \\
		& CRPS & 0.066 & 0.053 & 0.056 & {\color{red}0.181} & \textbf{0.049} \\
		\toprule
	\end{tabular}
\end{table}

Consider next the multi-fidelity set-up, with predictive metrics in Table \ref{tab: stress emulation error} and visualization in Figure \ref{fig: Blade prediction}. Again, the LOL-GP provides considerably improved predictions over existing methods in terms of RMSE and CRPS. Here, the KO and BKO models yield improvements over the standard GP (thus no negative transfer), whereas the BTLNet still experiences negative transfer. Figure \ref{fig: Blade rho estimate} (right) shows the estimated transfer function $\rho(\bm{x})$ for our model. In the bottom-right corner, where $x_2$ (pressure load on the suction side) is larger than $x_1$ (pressure load on the pressure side), we see that the LOL-GP assigns near-zero transfer. This again corroborates our earlier observation of local transfer from Figure \ref{fig: Blade MF stress solution}, where with $x_2 > x_1$, the simulated stress response can be considerably different over fidelity levels, and vice versa. By capturing this local transfer behavior in our surrogate model, the LOL-GP can facilitate more robust and effective transfer learning over existing methods.

\section{Conclusion}
\label{sec:con}
This paper presents a novel transfer learning model, called the LOL-GP, for probabilistic surrogate modeling of costly computer simulators. The idea is to transfer information from available data on related source systems for accurate predictions on the target system of interest. Existing transfer learning surrogate models may be prone to ``negative transfer'', in that its transfer of information may worse surrogate performance. To address this, the LOL-GP models for a key ``local transfer'' property that identifies regions where transfer is beneficial and regions where it can be detrimental. Such local transfer is expected in many scientific systems, where systems may behave similarly at certain parametrizations but differently at others. We derive an efficient Gibbs sampling algorithm for posterior predictive sampling on the LOL-GP, for both the multi-source and multi-fidelity settings. We then demonstrate the effectiveness of the LOL-GP over existing methods, in a suite of numerical experiments and an application for jet turbine design.

Given these promising results, there are many promising directions to explore for future research. One direction is experimental design: how should design points be selected on the target system to maximize surrogate performance given a limited computing budget? The development of novel experimental design methods for transfer learning can further improve performance for the LOL-GP, and we are investigating this as future work. {Another direction is the investigation of alternate activation functions within the LOL-GP for broader transfer learning applications.} Finally, we are exploring applications of transfer learning surrogates for accelerating decision-making in different scientific disciplines. One promising area is high-energy physics, where particle simulators are highly expensive (requiring thousands of CPU hours per run; see \cite{kumar2023inclusive}) but there exists a wealth of existing simulation data on related particle systems in the literature. We aim to investigate such areas for impactful applications. 


\newpage

\includepdf[pages=-]{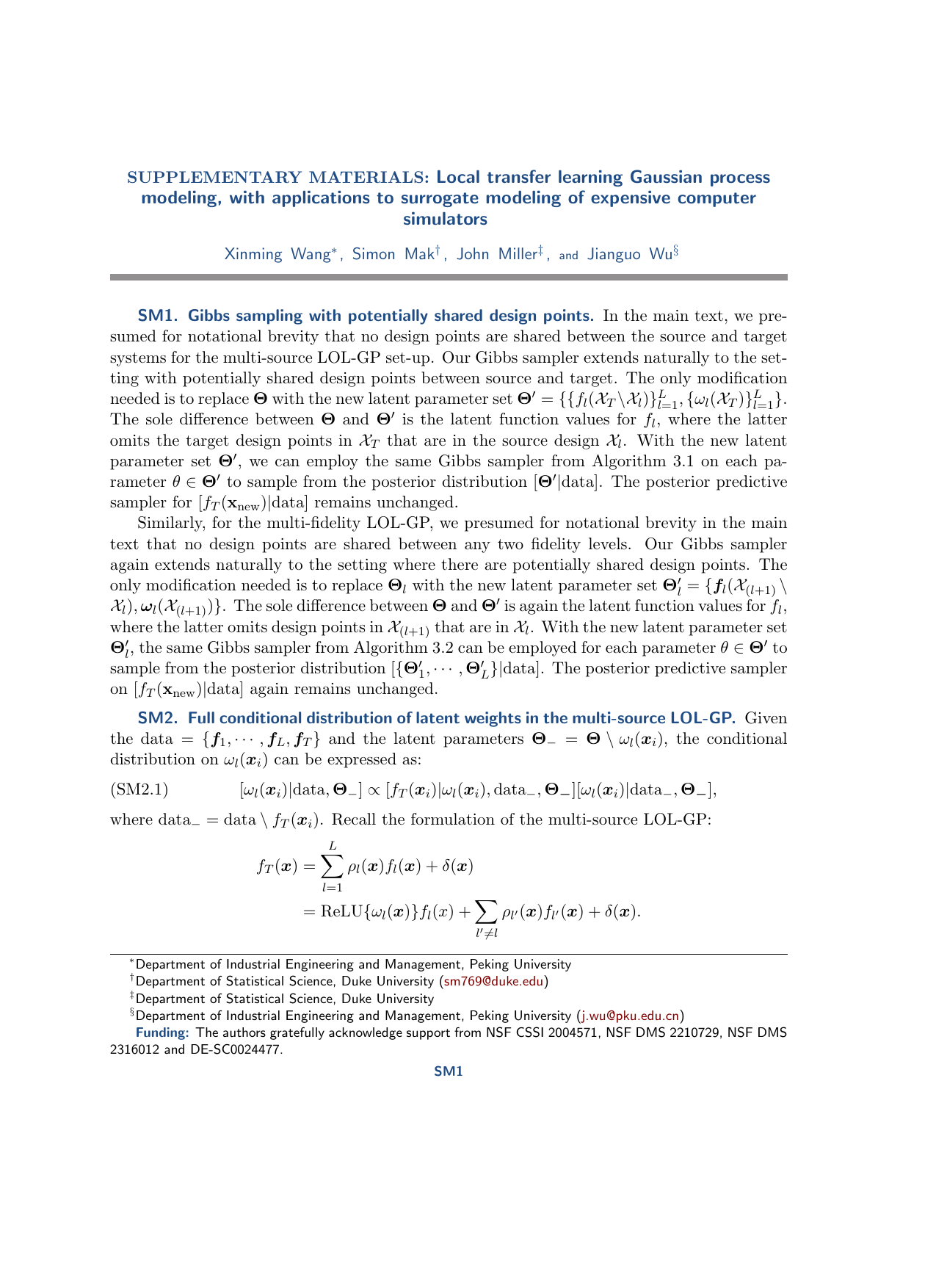}

\end{document}